%% file: example_paper.tex
\documentclass{article}

\usepackage{microtype}
\usepackage{graphicx}
\usepackage{subfigure}
\usepackage{booktabs} % for professional tables
\usepackage{hyperref}
\usepackage{url}
\usepackage{booktabs}
\usepackage{graphicx}
\usepackage{wrapfig}
\usepackage{xcolor}
\usepackage{latexsym}
\usepackage{graphicx}
\usepackage{tabularx}
\usepackage{multirow}
\usepackage{booktabs}
\usepackage{xcolor}
\usepackage{adjustbox}						
\usepackage{fancybox}
\usepackage{ulem}
\usepackage{wrapfig,lipsum,booktabs}
\usepackage{amsmath}
\usepackage{array}
\usepackage{makecell}
\usepackage{amsfonts}
\usepackage{footnote}
\definecolor{barca-blue}{RGB}{0, 76, 153}
\definecolor{barca-red}{RGB}{167, 0, 66}
\definecolor{lblue}{RGB}{13, 152, 255}
\definecolor{lred}{RGB}{255, 108, 108}
\definecolor{dblue}{RGB}{0, 112, 192}
\definecolor{dred}{RGB}{192, 0, 0}
\definecolor{dgray}{HTML}{7E7E7E}
\definecolor{tblue}{HTML}{174992}

\newcommand{\RRN}[1]{%
	\textup{\lowercase\expandafter{\it \romannumeral#1}}%
}

\usepackage{hyperref}

\usepackage[accepted]{icml2022}

\begin{document}

\twocolumn[
% \icmltitle{Sparse Attention-free Language Models}
% \icmltitle{Sparse all-MLP for Language Modeling}
\icmltitle{Efficient Language Modeling with Sparse all-MLP}
% It is OKAY to include author information, even for blind
% submissions: the style file will automatically remove it for you
% unless you've provided the [accepted] option to the icml2021
% package.

% List of affiliations: The first argument should be a (short)
% identifier you will use later to specify author affiliations
% Academic affiliations should list Department, University, City, Region, Country
% Industry affiliations should list Company, City, Region, Country

% You can specify symbols, otherwise they are numbered in order.
% Ideally, you should not use this facility. Affiliations will be numbered
% in order of appearance and this is the preferred way.
\icmlsetsymbol{equal}{*}

\begin{icmlauthorlist}
% Authors: Ping Yu, Mikel Artetxe, Myle Ott, Sam Shleifer, Hongyu Gong, Ves Stoyanov, Xian Li {artetxe, myleott, sshleifer, hygong, ves, xianl}@fb.com
\icmlauthor{Ping Yu}{equal,buf}
\icmlauthor{Mikel Artetxe}{fb}
\icmlauthor{Myle Ott}{fb}
\icmlauthor{Sam Shleifer}{fb}
\icmlauthor{Hongyu Gong}{fb}
\icmlauthor{Ves Stoyanov}{fb}
\icmlauthor{Xian Li}{fb}
\end{icmlauthorlist}

\icmlaffiliation{buf}{State University of New York at Buffalo, Buffalo, NY 14228}
\icmlaffiliation{fb}{Meta AI}
% \icmlaffiliation{ed}{School of Computation, University of Edenborrow, Edenborrow, United Kingdom}

\icmlcorrespondingauthor{Ping Yu}{pingyu@buffalo.edu}
\icmlcorrespondingauthor{Xian Li}{xianl@fb.com}

% You may provide any keywords that you
% find helpful for describing your paper; these are used to populate
% the "keywords" metadata in the PDF but will not be shown in the document
\icmlkeywords{Machine Learning, ICML}

\vskip 0.3in
]

% this must go after the closing bracket ] following \twocolumn[ ...

% This command actually creates the footnote in the first column
% listing the affiliations and the copyright notice.
% The command takes one argument, which is text to display at the start of the footnote.
% The \icmlEqualContribution command is standard text for equal contribution.
% Remove it (just {}) if you do not need this facility.

% \printAffiliationsAndNotice{}  % leave blank if no need to mention equal contribution
\printAffiliationsAndNotice{\icmlEqualContribution} % otherwise use the standard text.

\begin{abstract}
All-MLP architectures have attracted increasing interest as an alternative to attention-based models. In NLP, recent work like gMLP shows that all-MLPs can match Transformers in language modeling, but still lag behind in downstream tasks. In this work, we analyze the limitations of MLPs in expressiveness, and propose sparsely activated MLPs with mixture-of-experts (MoEs) in both feature and input (token) dimensions. Such sparse all-MLPs significantly increase model capacity and expressiveness while keeping the compute constant. We address critical challenges in incorporating conditional computation with two routing strategies. The proposed sparse all-MLP improves language modeling perplexity and obtains up to 2$\times$ improvement in training efficiency compared to both Transformer-based MoEs (GShard, Switch Transformer, Base Layers and HASH Layers) as well as dense Transformers and all-MLPs. Finally, we evaluate its zero-shot in-context learning performance on six downstream tasks, and find that it surpasses Transformer-based MoEs and dense Transformers.

% All-MLP architecture has attracted growing interest with competitive performance on par with Transformers. In NLP, recent work such as gMLP demonstrates that all-MLP can match Transformers in language modeling pretraining but still lags behind in downstream tasks performances. In this paper, we analyze the limitation of MLPs in expressiveness and propose sparsely activated MLPs with mixture-of-experts (MoEs) in both feature and input (token) dimensions. Such sparse all-MLP significantly increases model capacity and expressiveness while keeping the compute constant. We address critical challenges in incorporating conditional computation into all-MLPs with sparsely activated experts with two routing strategies. The proposed sparse all-MLP improves language modeling perplexity and obtains up to 2$\times$ improvement in training efficiency compared to both Transformer-based MoEs (such as Gshard, Switch Transformer, Base Layers, Hash Layers, etc.) as well as dense Transformers and MLPs under the same compute budget. Finally, we evaluate its zero-shot in-context learning performance on six downstream tasks ranging from commonsense reasoning to question answering and find it surpasses state-of-the-art sparse Transformer-based MoEs and closes the gap with dense Transformers (e.g. GPT-3). 
\end{abstract}

\input{content/1_introduction.tex}
\input{content/2_background.tex}

\input{content/3_methods}

\input{content/4_experiments}
\input{content/6_related_works}
\input{content/7_conclusion}
\nocite{langley00}

\bibliography{example_paper}
\bibliographystyle{icml2022}

%%%%%%%%%%%%%%%%%%%%%%%%%%%%%%%%%%%%%%%%%%%%%%%%%%%%%%%%%%%%%%%%%%%%%%%%%%%%%%%
%%%%%%%%%%%%%%%%%%%%%%%%%%%%%%%%%%%%%%%%%%%%%%%%%%%%%%%%%%%%%%%%%%%%%%%%%%%%%%%
% DELETE THIS PART. DO NOT PLACE CONTENT AFTER THE REFERENCES!
%%%%%%%%%%%%%%%%%%%%%%%%%%%%%%%%%%%%%%%%%%%%%%%%%%%%%%%%%%%%%%%%%%%%%%%%%%%%%%%
%%%%%%%%%%%%%%%%%%%%%%%%%%%%%%%%%%%%%%%%%%%%%%%%%%%%%%%%%%%%%%%%%%%%%%%%%%%%%%%

% \clearpage
\newpage

\appendix
% \onecolumn
\input{content/A_other_methods}

\end{document}

%% file: content/1_introduction.tex
\section{Introduction}
\label{sec:introduction}
Transformers have been the state-of-the-art architecture for natural language processing (NLP) tasks \citep{devlin2018bert,radford2018improving,raffel2019exploring,liu2019roberta,brown2020language}. Recently, architectures using solely MLPs (multi-layer perceptrons) have shown to be competitive, especially in computer vision tasks  \citep{mlp_mixer,liu2021pay,lee2021fnet,hou2021vision,lou2021sparse}. In the field of NLP, recent work, like gMLP \citep{gmlp}, shows that all-MLP architecture can match Transformers \citep{vaswani2017attention} in language modeling perplexity, but there is still a gap in downstream performance. 

In this paper, we aim to push the performance of all-MLPs in large-scale NLP pretraining \citep{devlin2018bert,radford2018improving,raffel2019exploring,liu2019roberta,brown2020language}. We analyze current all-MLP approaches, and find limitations in their expressiveness. Drawing on the recent success of Transformer-based Mixture-of-Experts (MoEs) \citep{gshard,switch,baselayer,hashlayer,alibaba}, we propose sparsely-activated all-MLPs as an alternative to address this limitations. Our proposed architecture -- \textit{sparse MLP} (sMLP for short)  replaces major dense blocks in gMLP \citep{gmlp} with sparse blocks. As a result, it improves the expressive power of gMLP while keeping the compute cost constant.

Specifically, we apply sparsity to the two fundamental operations that are common in NLP models: 
\begin{itemize}
\vspace{-2mm}
    \item \textbf{Hidden dimension operation:} We adopt the experts in feed-forward layers as used in recent Transformer-based MoEs like the Switch Transformer \citep{switch} and Base Layers \citep{baselayer}.
    % \vspace{-2mm}
    \item \textbf{Token-mixing operations:} This is implemented as self-attention in Transformers and the Spatial Gating Unit in gMLP. We observe that naive extensions in the Transformer architecture, such as turning the self-attention module into experts, performs poorly. We design a new MoE module, named sMoE, to address this challenge, which chunks the hidden representations of the input through the hidden dimension, sends chunked vectors to different expert, and performs spatial projection inside each expert.
\end{itemize}

% \vspace{-2mm}
We provide an in-depth analysis of the routing mechanisms in the sMoE module, where we point out that a less careful design can easily lead to information leaking when routing weights are learnt from future tokens. In the context of language modeling, the router makes use of future tokens during training, while at the inference time, the model needs to predict the tokens in an autoregressive fashion without looking into the future. To solve this problem, we propose two routing strategies : (1) \textbf{Deterministic Routing}: instead of learning the routing strategy, we directly send hidden vectors to each expert by cutting the hidden dimension. It has an indirect connection with multi-head attention mechanism \citep{vaswani2017attention}, where each expert serves as one head; (2) \textbf{Partial Prediction}: instead of learning the routing strategy from the whole sentence, we only use the first 20\% of tokens for learning the routing strategy and use it to predict the remaining 80\%. 

% At the beginning we used the standard MoE router for our sMoE module, which achieved fabulous performance. However, we notice that directly extend the standard routing method to token-wise operation will make the routing weights learn from future token information. That means, the router can make a "smart" decision because it could know all the information from the sentence. However, in the inference stage, the language modeling model need to predict the tokens aggressively then the router cannot make such a "smart" decision. Using the standard router is actually the upper bound of our method. In order to avoid such problems, we propose two improved methods: (1). deterministic routing: instead of learning the routing strategy, we directly send hidden vectors to each expert by cutting the hidden dimension. In this way, each expert serves as one head (multi-head mechanism \citep{vaswani2017attention}). (2). partial prediction: instead of learning the routing strategy from the whole sentence, we used the first 20\% of tokens for learning the routing strategy and make prediction for the rest 80\% of tokens; We trained a set of small models and a set of large models, experimenting on a small dataset and a large-scale dataset. The experiments results show that our sMoE-deterministic could converge faster, more stable, better perplexity and maintain a high inference speed. 

\begin{figure}
    \centering
    \includegraphics[width=0.75\linewidth]{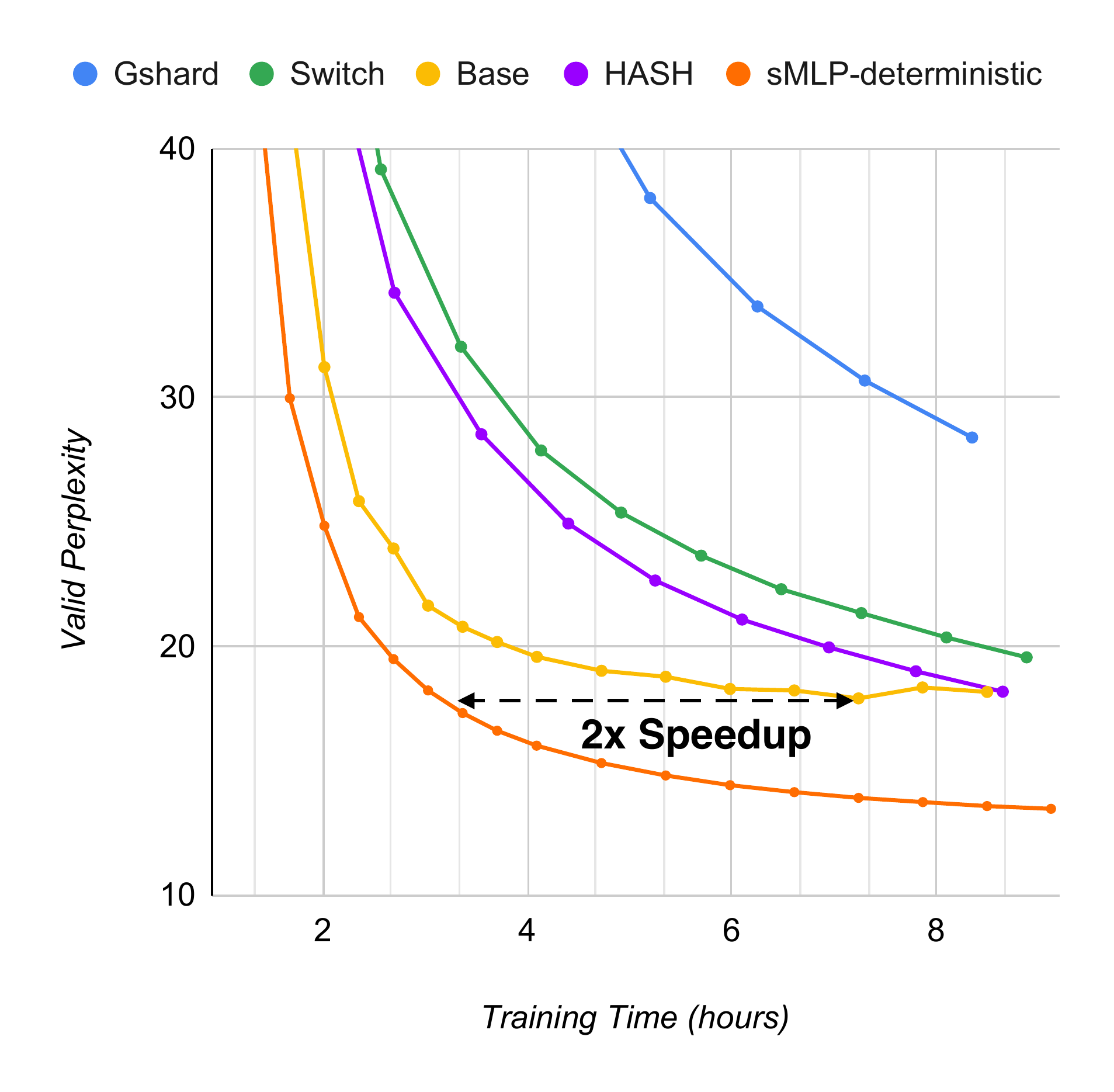}
    \vspace{-6mm}
    \caption{The proposed sparse all-MLP architecture (sMLP) achieves 2$\times$ training efficiency improvement compared to state-of-the-art sparse Transformer-based MoEs: Gshard\citep{gshard}, Switch Transformer \citep{switch}, Base Layers \citep{baselayer}, HASH Layers \citep{hashlayer}. Full comparison is provided in Fig.~\ref{fig:model_comparision}.} 
    % \vspace{-2mm}
    \label{fig:speed_up}
\end{figure}

To summarize, our contributions are as follows:
% \vspace{-4mm}
\begin{itemize}
% \vspace{-4mm}
    \item We propose a sparsely activated all-MLP architecture, named sMLP. To the best of our knowledge, this is the first NLP work combining all-MLP-based models with MoEs. We provide an in-depth analysis of why the MLP architecture lags behind Transformers in terms of expressiveness and identify two core challenges in turning MLPs into sparsely activated MoEs. The proposed sMLP architecture addresses these challenges with a novel sMoE module and two routing strategies. 
    % \vspace{-2mm}
    \item We empirically evaluate its performance on language modeling and compare with strong baselines of both sparse Transformer-based MoEs such as Gshard \citep{gshard}, Switch Transformer \citep{switch}, Base Layers \citep{baselayer} and HASH Layers \citep{hashlayer} as well as dense models including Transformer \citep{vaswani2017attention} and gMLP \citep{gmlp}. Our sMLP outperforms these models in terms of valid perplexities while obtaining up to 2 $\times$ improvement in pretraining speed with the same compute budget (shown in Fig.~\ref{fig:speed_up}). In addition, sMLP demonstrates good scalability where it still outperforms sparse Transformers counterparts when we scale sMLP to 10B parameters with a large pretraining corpus of 100B tokens.
    % \vspace{-2mm}
    \item Finally, we evaluate its zero-shot priming performance after pretraining on language modeling. Through head-to-head comparison, the proposed sparsely-activated all-MLP language model outperforms sparsely-activated Transformers on six downstream tasks ranging from natural language commonsense reasoning to QA tasks.
    Compared with dense Transformers such as GPT-3 \citep{gpt3}, our sMLP model achieves similar zero-shot performance despite being pretrained on a much smaller dataset (3 $\times$ smaller than the dataset used by GPT-3) with less computing resources.
\end{itemize}
% \paragraph{Sparse MLP model:} 
% We extend MLP-based model to sparse model architecture. To our best knowledge, this is the first work focusing on extending MLP-based model with MoE in the natural language processing field. We identified and addressed two core challenges in turning MLPs into sparsely activated mixture-of-experts, including proposing a novel expert module based on spatial linear layer and corresponding routing methods. 

% \vspace{-4mm}
% \paragraph{Applying MoE to new-dimension:}
% The standard MoE layer is pre-defined for feed-forward layers, which sends each tokens to experts. As a comparison, our sMoE layer send each hidden vector to experts, and perform spatial linear layer in each expert module. 

% \vspace{-4mm}
% \paragraph{In-depth discussion of transformer model structure and MLP model structure:} We analyzed and compared MLP-based model and transformer-based model. We found that the multi-head mechanism does not work on the MLP model. Finally, we explained the advantages of our model over dense MLP model. 

% \vspace{-4mm}
% \paragraph{Performance on language modeling task:}
% Our sMoE model can strongly outperform the current MoE models and MLP model on language modeling pre-training task. 

\begin{figure*}
    \centering
    \includegraphics[width=0.88\linewidth]{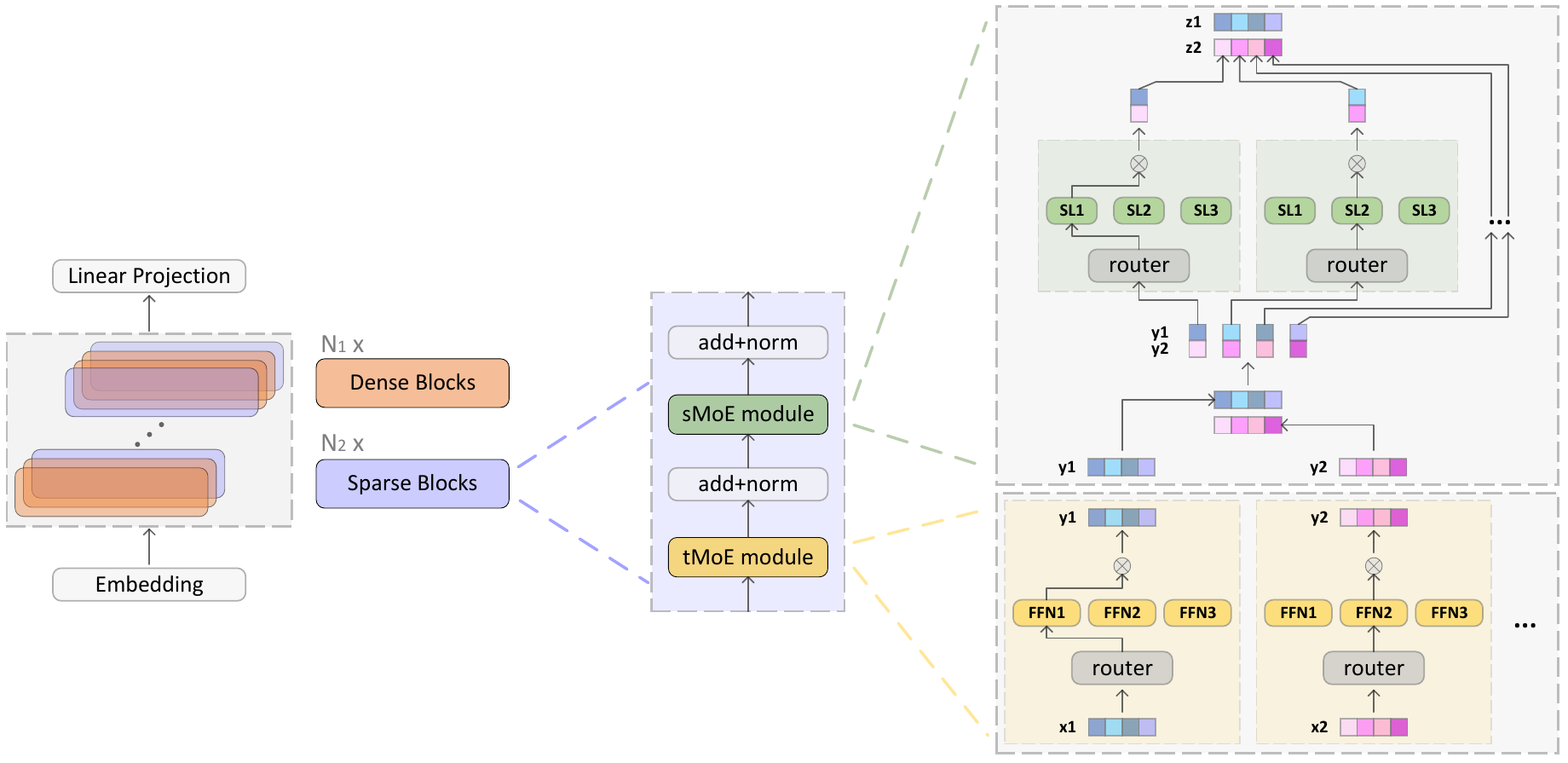}
    % \vspace{-2mm}
    \caption{\textbf{Illustration of sMLP Model Architecture} with $N_1$ Dense Blocks (gMLP layers) and $N_2$ Sparse Blocks. The arrangement of these blocks will be discussed in Section~\ref{sec:expe_setup}.
    Each sparse block contains a tMoE module and a sMoE module. The tMoE module sparsely activated FFN (feed-forward) and the sMoE module replaces the self attention in Transformers with sparse token-wise SL (spatial linear projection) operations.}
    % \vspace{-2mm}
    \label{fig:model_frames}
\end{figure*}

%% file: content/2_background.tex
% \vspace{-6mm}
\section{Background}
\label{sec:background}

\subsection{Token-wise Operations}

% In the computer vision field, Convolutional Neural Networks (CNNs) are the go-to model. Attention-based networks, such as the Vision Transformer, have also become popular. Recent findings from \citep{tolstikhin2021mlp,guo2021beyond,touvron2021resmlp,melas2021you} show that the architecture based exclusively on multi-layer perceptrons (MLPs) could achieve comparable or even superior results. 

% In the natural language processing field, transformers \citep{transformer2017} have become dominant architecture over the past few years. The self-attention module is an essential part of transformer architecture. Recently, a fully MLP model - gMLP \citep{gmlp} designs the Spatial Gating Unit to replace the self-attention module

%Previous work has two kinds of token-wise operations: (1). Self attention module in the Transformer \citep{vaswani2017attention} architecture; (2). Spatial Gating Unit in the gMLP \citep{gmlp} model. 
Transformers and all-MLPs perform token-mixing operations in different ways. The former uses self-attention \citep{vaswani2017attention}, while the latter uses Spatial Gating Unit \citep{gmlp}.

Given the input token representation $X \in \mathbb{R}^{T\times H}$, where $T$ represents the sequence length and $H$ represents the hidden dimension. We set $h$ to be the total head number. 

% if we set $h$ to be the head number, then the hidden dimension for input to a single head will be $d = H/h$.

% \xian{Either make this description apply to both self attention and Spatial Gating Unit or move it inside self attention. }
% Within the single head, the input is $X_i \in \mathbb{R}^{T \times d}$. 

\vspace{-2mm}
\paragraph{Self attention module}
The multi-head self attention module is a concatenation of token-mixing operations from each head:
\begin{equation}
\begin{aligned}
Y &= \operatorname{Concat}\left(M_{1}, \ldots, M_{\mathrm{h}}\right) \\
\text { with }M_{\mathrm{i}} &= \operatorname{Softmax}(XW_i^Q\ XW_i^K) XW_i^V
\end{aligned}
\label{eq:self_attention}
\end{equation}
where $Y$ is the output, $W_i^Q, W_i^K, W_i^V \in \mathbb{R}^{H \times d}$ are projection matrices. $d$ is the hidden dimension for input to a single head $d = H/h$.

%Due to the reduced dimension of each head, the total parameters is similar to that of single-head attention with full dimensionality. 
% At the same time, the total amount of parameters will not be changes when the sequence length increase.

% Each self attention module contains $h$ $W_Q, W_K and W_V$. Thus, the total number of parameters is $3*h*d = 3*h*(H/h) = 3*H$. It can be seen that the parameter number is only related to the hidden dimension. As the number of heads increases, the parameter number will not change. 

\vspace{-2mm}
\paragraph{Spatial Gating Unit}
gMLP \citep{gmlp} designs the Spatial Gating Unit to replace the self attention module. Within each head, 
\begin{equation}
    Y = W_sX
    \label{eq:gmlp}
\end{equation}
, where $W_s \in \mathbb{R}^{T \times T}$. Note that, $W_s$ corresponds to the \text{attention score} instead of $W^V$ in Equation~(\ref{eq:self_attention}). 

% Each Spatial Gating Unit contains $h$ projection matrices $W_s$. Then the total number of parameters is $h*T*T$. Compared to Transformer self attention, a shortcoming of the Spatial Gating Unit is that the number of parameters increases linearly with the number of heads $h$ and quadratically with sequence length $T$.  

\subsection{Sparse Expert Models}
\label{sec:backgroud_moe}

% Large-scale pretraining has been demonstrating an effective path towards flexible and powerful neural language models. It achieved tremendous success across several fields, especially neural language processing. The mainstream methods \citep{devlin2018bert,radford2018improving,raffel2019exploring,liu2019roberta,brown2020language} gradually expend the model size of a densely-activated Transformer \citep{vaswani2017attention} to increase the model capacity and representation ability. However, training and serving such models are expensive. This is partially because these deep networks are typically "dense" - every sample is proceed using all of the parameters - thus, scale comes at a high computational cost. Inspired by the success of model scale but seeking greater computational efficiency, people begin to explore the sparse model architecture. 

Sparse expert models provide a highly efficient way to scale neural network training. Compared to the standard dense models that require extremely high computational cost in training, sparse expert models are shown to be able to deal with a much larger scale of data and converge significantly faster. In sparse models, model weights are distributed to different workers according to the MoE mechanism \citep{gshard,switch,baselayer,hashlayer,alibaba}.
%The transformer~\citep{transformer2017} structure contains two parts: token-wise operations (self attention module) and hidden representation operations (feed forward module). 
The standard MoE was initially being designed for the feed-forward module, which leverages a routing algorithm to dispatch each token to $k$ sub feed-forward modules. These sub feed-forward modules are allocated independently on different devices, which are also called {\em experts}. MoE allows processing input tokens in parallel specified by a gating function to achieve maximum computational efficiency. 

% \vspace{-4mm}
\paragraph{MoE Routing}
Let $\{E_i(x)\}_{i=1}^N$ be a set of $N$ experts for a token representation $x$. 
\citet{shazeer2017outrageously} proposed a MoE layer that learns to dispatch the token representation to the best determined top-$k$ experts, selected from a set $\{E_i(x)\}_{i=1}^N$ of $N$ experts.
The router variable $W_r$ produces logits $h(x) = x \cdot W_r$ which are normalized via a softmax distribution over the available $N$ experts at that layer. The gate-value for expert $i$ is given by,
\begin{equation}
    p_i(x) = \frac{e^{h(x)_i}}{\sum_{j=1}^N e^{h(x)_j}}
\label{eq:gating}
\end{equation}
%\vspace{-2mm}
The top-$k$ gate values are selected for routing the token $x$.
If $\mathcal{T}$ is the set of selected top-$k$ indices then the output computation of the layer is the linearly weighted combination of each expert's computation on the token by the gate value,
\begin{equation}
    y = \sum_{i \in \mathcal{T}} p_i(x) E_i(x) 
\end{equation}
% \vspace{-2mm}
Given that each token will be sent to different devices (experts), unbalanced loading adds heavy affects the efficiency of the model. The Switch Transformer \citep{switch} designed a differential load balancing loss to reduce communication latency of sending tokens to different devices.  
Base Layers \citep{baselayer} further simplify the framework by eliminating the balancing loss, adapting the balancing assignment algorithm \citep{bertsekas1992auction} to send tokens from each batch to different devices equally. Instead of learning the routing weight $W_r$, HASH Layers \citep{hashlayer} use a random hashing function as the routing gate. In order to distinguish this standard Transformer-MoE routing \citep{gshard,switch,baselayer,hashlayer,alibaba} from ours, we call this method tMoE.

% \mikel{maybe clarify where this is coming from (transformer-MoE I guess), I couldn't make sense of it at first.}

%% file: content/3_methods.tex
\section{Methods}
\label{sec:methods}

% In Appendix.\ref{sec:challenges}, we try to extend current gMLP model to MoE structure directly. This approach is a straightforward extension, but the performance is not good. Through our analysis, it has serious drawbacks: the performance will decrease as the increase of number of devices. 
A straightforward extension of gMLP to MoE structure hurts performance, which we provide a detailed analysis in Appendix~\ref{sec:challenges}. Instead of sending tokens to different experts, in Section \ref{sec:sparse_mlp} we propose a novel sMoE module to send hidden vectors to different experts. We further propose two corresponding routing strategies in Section \ref{sec:deterministic routing} and Section \ref{sec:partial_prediction}. 

% \mikel{general comment: I'd use Section X in capital throughout the paper; and either Figure X or Fig. X, but not Fig.X}

% Our goal is to extend an fully MLP architecture with mixture of experts. A straightforward way to achieve that is directly applying token-wise routing to Spatial Gating Unit. But we found that this direct approach will make the result worse. More details can be found in Appendix \ref{sec:challenges}.

\subsection{Sparsely-activated all-MLP}
\label{sec:sparse_mlp}
 
The overall architecture of sMLP is illustrated in Fig.~\ref{fig:model_frames}. Our sMLP model contains $N_1$ dense blocks and $N_2$ sparse blocks. Both $N_1$ and $N_2$ are hyper-parameters. In each sparse block, it contains two modules: 
% \begin{itemize}
%     \item  tMoE module: we adopt the MoE from Base Layers \citep{baselayer} to replace FFN module.
%     \item sMoE module: we design this sMoE module to replace the self attention module in the Transformer \citep{vaswani2017attention} and the Spatial Gating Unit in gMLP \citep{gmlp}.
% \end{itemize}
$(\RRN{1})$ \textbf{tMoE module}: we adopt the MoE from Base Layers \citep{baselayer} to replace the FFN module in dense Transformers \citep{vaswani2017attention}; 
% \mikel{This is a more general point but I find the terminology in the paper a bit confusing. What you are calling an FFN here is also a MLP technically, but you seem to use MLP to refer to attention-less models more broadly which I find weird, but then in the title you specify all-MLP which seems a better term for me but it's not used consistently throughout the paper...}
$(\RRN{2})$ \textbf{sMoE module}: we design this sMoE module to replace the self attention module in the Transformer \citep{vaswani2017attention} and the Spatial Gating Unit in gMLP \citep{gmlp};

% \subsubsection{Routing methods}
% \subsubsection{Method}
% \label{sec: routing_methods}
% \xian{nit: is it possible to make this figure at top-right of this page?}
Both tMoE and sMoE blocks contain two elements: % (1). \textbf{a Gating Function}: decide the strategy for sending input to expert modules; (2). \textbf{Expert Modules}: each devices (e.g., GPUs) contains expert modules. The parameters inside of each expert module are different since gradient doesn't need to aggregate across devices; 
% \vspace{-4mm}
\paragraph{Expert Modules} These are the modules that process the input.
The tMoE module contains an FFN in each expert. For our sMoE module, each expert contains the Spatial Gating Unit as is shown in Fig.~\ref{fig:gmlp_moe} (\textbf{Right}) in Appendix~\ref{sec:challenges}. 

% \vspace{-4mm}
\paragraph{Gating Function}
% \paragraph{Sparse Routing.} 
This is the module that decides which expert should process each part of the input. tMoE uses standard token-wise routing (described in Section~\ref{sec:backgroud_moe}). However, naively applying token-wise routing to gMLP is flawed, as it sends tokens from the same sentence to different experts. Tokens can only attend to previous tokens in the same device, which means that as the number of experts increases, the number of previous tokens that each word can attend will decrease. Refer to Appendix~\ref{sec:challenges} for more details. For that reason, we have to design a distinct routing method to extend the MoE structure to the feature dimension. 

% \mikel{I rewrote this whole paragraph, please check that it makes sense}
% The gMLP MoE in Appendix~\ref{sec:challenges} directly used the token-wise router (the standard way in MoE), which will cause a severe problem: tokens from the same sentence will be sent to different devices. Tokens can only attend to previous tokens in the same device, which means that as the number of experts (devices) increases, the number of previous tokens that each word can attend will decrease. Unlike gMLP MoE, we have to design a distinct routing method to extend the MoE structure to the feature dimension.

%To make our description clearer, we use a specific example to illustrate how to send different hidden vectors to different experts. 

\begin{figure}[!t]
   \begin{minipage}{0.24\textwidth}
     \centering
     \includegraphics[width=.95\linewidth]{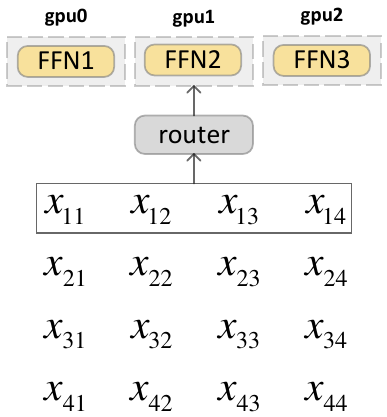}
    %  \caption{\textbf{tMoE gating:} Sending 4 tokens to 3 experts.}\label{fig:moe}
   \end{minipage}\hfill
   \begin{minipage}{0.23\textwidth}
     \centering
     \includegraphics[width=.85\linewidth]{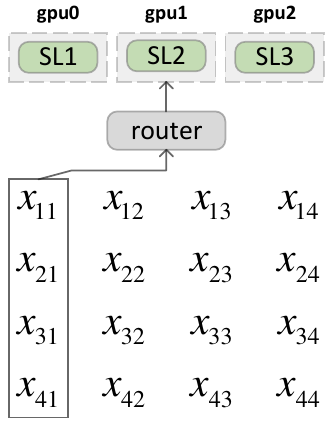}
    %  \caption{\textbf{sMoE gating}: Sending 4 hidden vectors to 3 experts.}\label{fig:smoe}
    %  \vspace{-2mm}
   \end{minipage}
   \caption{\textbf{Left}: tMoE Gating: sending 4 tokens to 3 experts; \textbf{Right}: sMoE Gating: sending 4 hidden vectors to 3 experts.}
   \label{fig:moe_smoe}
\end{figure}

Fig.~\ref{fig:moe_smoe} (\textbf{Left}) shows an example of the gating function from existing Transformer-based MoEs \citep{gshard,switch,baselayer,hashlayer}. $x_{ij}$ denotes the value of the $j_{th}$ hidden dimension in the $i_{th}$ tokens. The tMoE sends these four tokens to these three experts at the FFN layer using a learnt gating function described in Eq.(\ref{eq:gating}) parameterized by $W_r \in \mathbb{R}^{4 \times 3}$.
% \footnote{Note that this gating way is the standard gating method used in all previous MoE models. In order to distinguish this method from ours, we call the previous gating method tMoE.}; 
% where each expert module is a feed-forward module. 
% The main problem in MoE is how to route these four tokens to three different experts. \citep{gshard,switch,baselayer,alibaba} adopt a learnable gating method described in Eq.\ref{eq:gating}. In this example, it is to learn a router variable $W_r \in \mathbb{R}^{4 \times 3}$.

Unlike these existing MoEs, in sparse all-MLP architecture we propose to chunk hidden representation along the hidden dimension and send chunked vectors to different experts, as is shown in Fig.~\ref{fig:moe_smoe} (\textbf{Right}). 
% In this example, we cut this sentence to 4 hidden vectors by splitting hidden dimension.
We next discuss the design of the gating function for this new routing mechanism. % In order to send each hidden vector to different experts, we also need a gating function to learn how to assign different hidden vectors to different experts.

% \vspace{-2mm}
\subsection{Routing in Feature Space}
\label{subsec:routing_challenges}
% We first try to use the previous tMoE gating method to learn how to allocate the hidden vector. We found that this approach inevitably needs to obtain information from future tokens when learning routing gate $W_r$. Moreover, 
Compared to routing tokens, routing hidden dimensions faces a unique challenge in autoregressive models, with information leaking from looking ahead at future tokens if done naively (more details can be found in Appendix~\ref{sec:normal_gating}.). Furthermore, unlike Transformers-based MoEs with self-attention, appropriate masking cannot be directly applied to prevent information leaking. Due to the aforementioned problems, we cannot adopt existing routing methods in Transformers-based MoEs for language modeling. We propose and compare the following two solutions: deterministic routing and partial prediction. 

% \begin{figure}
%     \centering
%     \includegraphics[width=.5\linewidth]{old_img/normal_gating.png}
%     \caption{Apply original gating method on the sMoE}
%     \label{fig:original_gating}
% \end{figure}

\subsubsection{Deterministic Routing}
\label{sec:deterministic routing}

In order to decide where to send different hidden vectors, we need to learn a gating weight $W_r \in \mathbb{R}^{T\times N}$. This gating weight inevitably needs to multiply the hidden vector $v$ to obtain the routing probability $p_i(v)$. This vector is one specific hidden dimension of the whole sentence, including future tokens.

In order to prevent the gating weights from exploiting future token information, which is not available at inference time, our first approach removes $W_r$ directly. Instead, we chunk the vector in a hidden dimension and send hidden vectors to experts deterministically. In this case, if we want to send the hidden vector $V \in \mathbb{R}^{H\times T}$ to $N$ expert modules, we will chuck $h(v)_i \in \mathbb{R}^{H/N\ \times\ T}$ from hidden vector $V$. If $i$ equals to $0$, the gating will send the first to $H/N$ hidden vectors to $E_1(v)$. 

% \mikel{I find this notation awkward}
% \begin{equation}
%     y = \sum_{i \in \mathcal{H}} h(v)_i E_i(v) 
% \end{equation}

% This method is similar to the multi-head mechanism in the self attention module. The self attention module chucks hidden dimensions as different heads to calculate the attention score. However, this sMoE method chucks hidden dimensions to send different hidden vectors to an expert module. 

Unlike the Spatial Gating Unit, our attention-free sMoE-deterministic module splits the hidden dimensions among multiple experts. In each expert module, we insert one Spatial Gating Unit, similar to multi-head attention, where each expert serves as one head. The total number of parameters is $N*T*T$ where  $N$ is the number of experts/heads. Moreover, with each head being fitted to one device, the computation is parallel. 

% \xian{This paragraph should be moved to Methods, i.e. Sec 3.1. The current section is about background. }

\subsubsection{Partial Prediction}
\label{sec:partial_prediction}

Another way to prevent the gating weight $W_r$ from using future token information is partial prediction. Instead of predicting all the tokens in the sentence, we condition on the first $20\%$ of tokens, which are used to decide the routing, and predict the remaining $80\%$ of tokens. 

Given the input token representation $X \in \mathbb{R}^{T\times H}$, we divide it into two parts in the token dimension: the first $20\%$ of tokens $X_1 \in \mathbb{R}^{0.2T \times H}$, and the remaining $80\%$ of tokens $X_2 \in \mathbb{R}^{0.8T \times  H}$. Instead of training the language model on the whole sequence length $T$, we only train it to predict $X_2$. We use $X_1$ to learn the gating weights $W_r$. In sMoE, we transpose theinput $X_1$ to $V_1 \in \mathbb{R}^{H \times 0.2T}$, input $X_2$ to $V_2 \in \mathbb{R}^{H\ \times\ 0.2T}$. The router variable $W_r \in \mathbb{R}^{0.2T \times N}$ produces $h(V_1) = V_1 * W_r$. $V_1$ contains $H$ hidden vectors $v_i \in \mathbb{R}^{0.2T}$. The probability for sending $i_{th}$ hidden vector is learned by
\begin{equation}
    p_i(v) = \frac{e^{h(v)_i}}{\sum_j^N e^{h(v)_j}}
\end{equation}
% using Eq. \ref{eq:smoe_gating}. 

Different from the previous method, after learning the probability $p_i(v)$, this partial prediction method sends hidden vectors $v_i \in \mathbb{R}^{0.8T}$ from $V_2$ to the expert module instead of hidden vectors from $V_1$.

% We first try to use the previous MoE gating method to learn how to allocate the hidden vector. The MoE layer take token representation $X \in \mathcal{R}^{T*H}$, where $T$ represents the sequence length and $H$ represents the hidden dimension. In our sMoE, we transpose input to the hidden vector $V \in \mathcal{R}^{H*T}$. Our gating function routs this to the best top-$k$ experts, selected from a set $\{E_i(v)\}_{i=1}^N$ of $N$ experts. The router variable $W_r \in \mathcal{R}^{T * N}$ produces $h(V) = V W_r$ which are normalized via a softmax distribution over the available $N$ experts at that layer. $V$ contains $H$ hidden vector $v_i\in \mathcal{R}^{T}$. The probability for sending $i_{th}$ hidden vector is given by, 
% \begin{equation}
%     p_i(v) = \frac{e^{h(v)_i}}{\sum_j^N e^{h(v)_j}}
% \label{eq:smoe_gating}
% \end{equation}
% The top-$k$ gate values are selected for routing the hidden vector $v$. If $\mathcal{T}$ is the set of selected top-$k$ indices then the output computation of the layer is the linearly weighted combination of each expert's computation on the token by the gate value,
% \begin{equation}
%     y = \sum_{i \in \mathcal{N}} p_i(v) E_i(v) 
% \end{equation}

%% file: content/4_experiments.tex
\section{Experiments and Results}

\subsection{Experimental Setup}
\label{sec:expe_setup}

% \paragraph{Task} We evaluate the proposed architecture on language modeling task, as recent work such as GPT3 \citep{gpt3} demonstrates the language models pretraining is capable of performing various downstream tasks with few labelled examples, and the performance from scaling up model size has not been saturated yet.  % The previous MoE models \citep{gshard,switch,baselayer,hashlayer,alibaba} all focused on language modeling.

% \vspace{-2mm}
\paragraph{Baselines} 
% Our model extends gMLP \citep{gmlp} to a sparse version. A sparse block contains a tMoE module from BASE Layers \citep{baselayer}. In addition, we adapted the balanced assignment routing method for our sMoE implementation. Thus, the most direct baselines for our model should be gMLP \citep{gmlp} and BASE Layers \citep{baselayer}. We compared it with the current state-of-the-art dense (Transformer \citep{vaswani2017attention}, gMLP \citep{gmlp}) and sparse models (Gshard \citep{gshard}, BASE Layers \citep{baselayer}, HASH layers \citep{hashlayer}) as listed in Table \ref{tab:baselines} in Appendix \ref{sec:app_baselines}.

We compare to strong baselines consisting of state-of-the-art dense and sparse models, as is summarized in Table~\ref{tab:baselines}. We train all our baselines (except GPT3 from paper) and our model in PyTorch \citep{pytorch} using FAIRSEQ \citep{fairseq}.

\begin{table}
\centering
\scalebox{0.8}{
\begin{tabular}{l| c c}
    \toprule
        Baselines & Model  \\
        \midrule
         Dense Baselines  & \makecell{Transformer \citep{vaswani2017attention}\\  gMLP \citep{gmlp}} \\
         \midrule
          Sparse Baselines & \makecell{ Gshard \citep{gshard} \\ Switch Transformer \citep{switch} \\ Base Layers \citep{baselayer}\\ HASH Layers \citep{hashlayer} } \\
         \bottomrule
    \end{tabular}}
    \caption{\textbf{Baselines}: we compare our sMLP with  dense Transformers and MLPs as well as sparse Transformer-based MoEs.}\label{tab:baselines}
\end{table}

\begin{table}
\centering
\scalebox{0.65}{
\begin{tabular}{l| c | c |c}
    \toprule
        Experiments & FLOPs & Model Size & Datasets \\
        \midrule
         \makecell{Main\\Experiments}  & 0.8T & 3.50B & 1/10 data from RoBERTa  dataset \cite{liu2019roberta} \\
         \midrule
          \makecell{Scalability\\Experiments} & 2.0T & 9.41B &  \makecell{$\sim$ 100B tokens. RoBERTa and the English subset \\ of the CC100 corpus   \citep{conneau2019unsupervised}} \\
         \bottomrule
    \end{tabular}}
    \caption{\textbf{Experimental Settings}: In Section~\ref{sec:small_model_results}, we run our small model (3.50B parameters) with FLOPs per batch tokens 0.8T on the small dataset (1/10 of the RoBERTa dataset); In Section~\ref{sec:large_scale}, we run our large model (9.41B parameters) with FLOPs per batch tokens 2.0T on the whole RoBERTa dataset plus CC100 corpus.}\label{tab:datasets}
\end{table}

\begin{table}[t!]
\centering
\scalebox{0.8}{
    \begin{tabular}{l|c c c}
    \toprule
        Modules & heads & params.(M) & FLOPs (G) \\
        \midrule
         Self-attention & 16 & 4.198  & 12.885  \\ 
         Self-attention & 1 & 4.198  & 12.885  \\ 
         Spatial Gating Unit & 16 & 16.798  & 4.316 \\
         Spatial Gating Unit & 1 & 1.054  & 4.316 \\
         s-MoE & 1 & 1.058  & 4.387 \\
         \bottomrule
    \end{tabular}}
    \caption{\textbf{Token-wise Operations Comparison}: we compare our method with the previous two token-wise operations regarding parameters and FLOPs by each operation.}
    \label{tab:token_wise}
\end{table}

\begin{table*}[t!]
\centering
\scalebox{0.8}{
\begin{tabular}{l| c c c c c c c c}
    \toprule
        Models  & \makecell{Model Size \\ (Parameters)}&\makecell{Quality after 25k steps \\(Perplexity $\downarrow$)} &  \makecell{Time to Quality\\ (hours $\downarrow$)} & \makecell{Speed\\ (world per second $\uparrow$)} \\
        \midrule
        Gshard \citep{gshard} & 3.48B & 19.26 & 27.54 & 136k \\
        Switch Transformer \citep{switch} & 3.48B & 15.31 & 20.67 & 185k \\
        Base Layers \citep{baselayer} & 3.60B &18.16 & 21.13 & 178k \\
        HASH Layers \citep{hashlayer} & 3.51B &13.57 & 21.08 & 176k \\
        \midrule
        sMLP -- deterministic (our model)& 3.50B & \textbf{13.46} & \textbf{20.41} & 192k\\
         \bottomrule
    \end{tabular}}
    \scalebox{0.8}{
    \begin{tabular}{l|c c c c c c c c c}
    \toprule
        Models  &\makecell{Model Size \\ (Parameters)} & \makecell{Quality after 100k steps \\(Perplexity $\downarrow$)} &  \makecell{Time to Quality\\ (hours $\downarrow$)} & \makecell{Speed\\ (world per second $\uparrow$ )}\\
        \midrule
        Gshard \citep{gshard} & 9.03B & 14.5 & 198 & 115k\\
        Switch Transformer \citep{switch} &  9.03B & 10.45 & 120 & 139k\\
        Base Layers \citep{baselayer} & 10.31B &20.36 \footnotesize{(divergence)} & 115.13 & 135k\\
        HASH Layers \citep{hashlayer} & 10.29B &16.69 & \textbf{100} & 141k\\
        \midrule
        Switch Transformer - Enlarge (FLOPs 2.3T) & 10.31B & 10.09 & 140 & 126k \\
        \midrule
        sMLP -- deterministic (our model)& 9.41B & \textbf{9.84} & 112 & 144k\\
         \bottomrule
    \end{tabular}}
    \caption{Head-to-head comparison measures per step and per time benefits of the sMLP (our model) over baselines. We report perplexity (lower better) and time to reach (lower better) as quality and training efficiency measures. All models are trained with the same amount of computation and hardware. In order to test parameter-matched models, we train an enlarged Switch Transformer with 2.0T FLOPs. \textbf{Top}: Small models (FLOPs 0.8T). \textbf{Bottom}: Scaled-up models (FLOPs 2.0T). More model details are described in Table~\ref{tab:bigger_size_model_settings}.}\label{tab:pre-training-results}
\end{table*}

% \vspace{-2mm}

\begin{figure}
    \centering
    \includegraphics[width=0.75\linewidth]{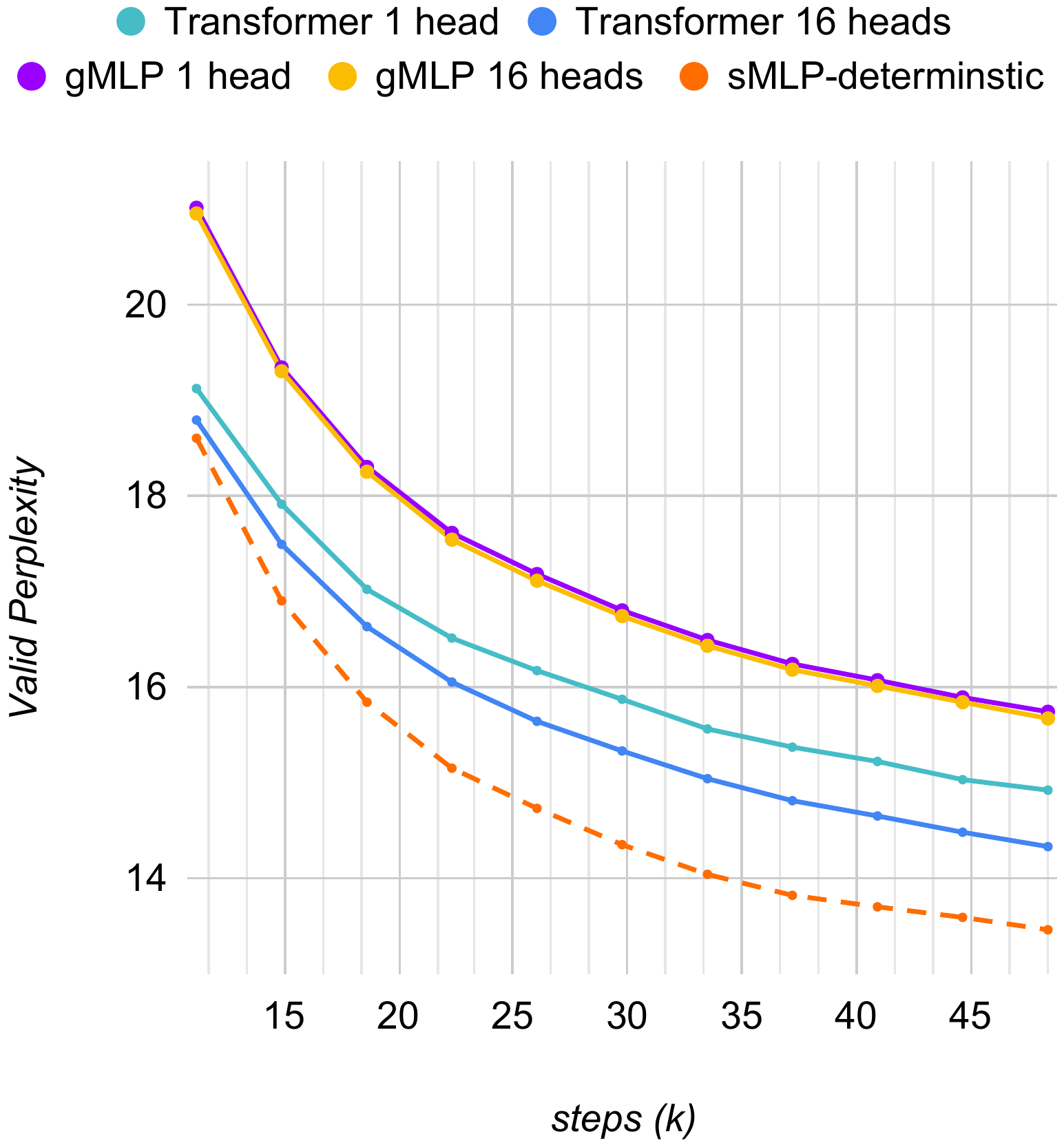}
    \vspace{-2mm}
    \caption{\textbf{Comparison with Dense Models}: We compare our model (the orange dashed line) with dense models (solid lines), including both Transformer \citep{transformer2017} and gMLP \citep{gmlp}, with different heads. Our model (sMLP) only has one head activated on each device.}
    \label{fig:dense_comparison}
\end{figure}

\vspace{-2mm}
\paragraph{Experimental Settings and Dataset}

A sparsely activated model splits \textit{unique} weights on different devices. Therefore, the weights of the model increase with the number of devices, all while maintaining a manageable memory and computational
footprint on each device. Previous work \citep{gshard,switch,alibaba} controls the floating point operations (FLOPs) per example for model comparison. We utilize the fvcore library \footnote{FLOPs are calculated for the forward pass in \url{https://github.com/facebookresearch/fvcore}} to measure the floating point operations (FLOPs) per batch tokens during evaluation for model comparison. Since all models contain exactly the same tokens in each batch, we actually controls FLOPs per token.

Our experimental settings are summarized in Table~\ref{tab:datasets}. $(\RRN{1})$ we compare to all baselines in a regular setting. In order to control for FLOPs, we change the number of layers in the model. More concretely, since Gshard \citep{gshard} and Switch Transformer \citep{switch} have larger FLOPs values for the same model structure, we reduce their number of layers while keeping the same hidden dimension to achieve the same FLOPs. $(\RRN{2})$ we also conduct experiments at larger scale and data size to test its scalability. We use pretraining datasets from RoBERTa \citep{liu2019roberta} and the English portion of CC100 \citep{conneau2019unsupervised}. More details can be found in Appendix~\ref{sec:appendix_datasets}. 

% \mikel{how did you get these FLOP estimates? we estimated 0.4 ZFLOPs for training a 125M GPT-3 model, which is 500,000,000 times more than what you estimate here. are these FLOPs per sequece or something like that? please clarify and double check your estimates are sounds}

% In order to have a fair model comparison, researchers generally control the same model size (parameter number) for all the models. 
% However, for a sparse model, not all the parameters will be activated, researchers \citep{gshard,switch,alibaba} controls FLOPs \footnote{FLOPs is measured by fvcores library: \url{https://github.com/facebookresearch/fvcore}} (which measures the number of floating-point operations) for model comparison. 

%\paragraph{Dataset} 
%To comprehensively investigate the sMoE model, we conduct experiments on different settings as shown in Table \ref{tab:datasets}. RoBERTa contains five English-language corpora of varying sizes and domains. The details can be found in Appendix \ref{sec:appendix_datasets}. 

% \begin{figure}[t!]
%     \centering
%     \includegraphics[width=0.75\linewidth]{img/ppl_steps.pdf}
%     \vspace{-2mm}
% \end{figure}
% \begin{figure}
%     \centering
%     \includegraphics[width=0.75\linewidth]{img/ppl_time_crop.pdf}
%     % \vspace{-3mm}
%     \caption{\textbf{Top}: Valid Perplexity v.s. Steps; \textbf{Bottom}: Valid Perplexity v.s. Training Time. We compare our two methods (sMLP-deterministic and sMLP-partial) with current sparse models with 0.8T FLOPs.}
%     \label{fig:model_comparision}
% \end{figure}

\begin{figure}[t!]
    \centering
    \includegraphics[width=0.7\linewidth]{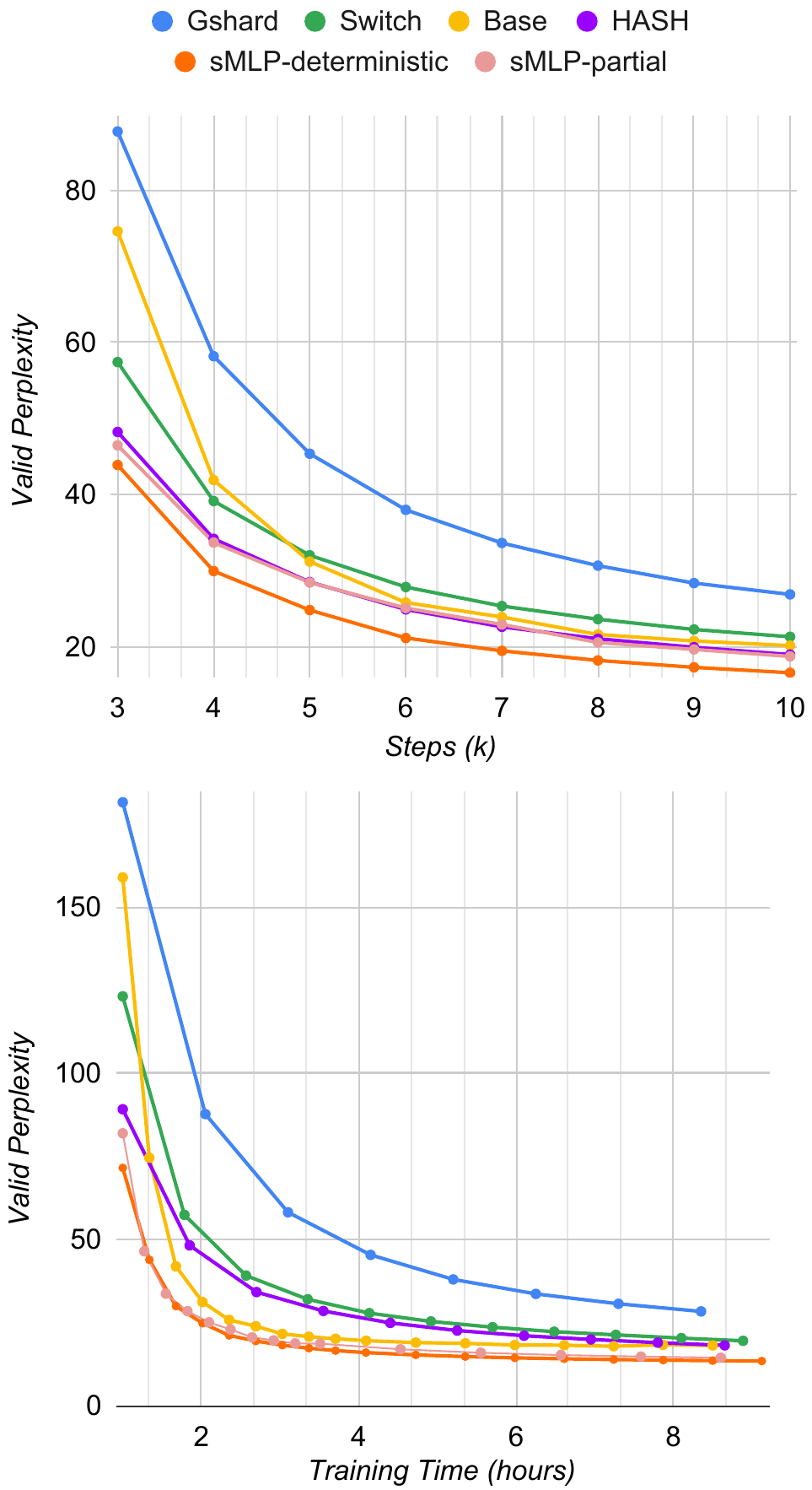}
    \vspace{-2mm}
    \caption{\textbf{Top}: Valid Perplexity v.s. Training Steps; \textbf{Bottom}: Valid Perplexity v.s. Training Time. We compare our two methods (sMLP-deterministic and sMLP-partial) with sparse Transformer MoEs with 0.8T FLOPs.}
    % \vspace{-2mm}
    \label{fig:model_comparision}
\end{figure}

\vspace{-2mm}
\paragraph{Metrics} (1). \textbf{Language modeling perplexity:} We report validation perplexity as the indicator of language modeling performance. (2). \textbf{Training efficiency:} For measuring training speed and efficiency, we compare by both number of updates and word-per-second (WPS). (3). \textbf{Accuracy:} we report accuracy on downstream tasks to measure performance of zero-shot priming. % \textbf{Training Speed:} A good model needs to run fast, we adopt the word per second to measure the training speed during the training stage. (4). \textbf{Accuracy:} Excellent language modeling results are not necessarily translating to downstream task performance improvement. We conduct zero-shot evaluations on downstream understanding tasks using our pretrained checkpoints. These tasks utilize accuracy as the metric.

\begin{table*}[t!]
    \centering
    \scalebox{0.8}{
    \begin{tabular}{c|c c c c c c c c c}
        \toprule
         Models & FLOPs (T) & \makecell{Model\\ Size (M)} & \makecell{Embedding\\ Dim} & \makecell{Hidden\\ Dim} & \makecell{Sparse \\Layers} & \makecell{Dense\\ Layers} & \makecell{Num of \\Heads} & \makecell{Num of\\ Experts} & \makecell{Num of \\ GPUs} \\
         \midrule
         Gshard  & 2.0  & 9.03  & 2048  & 8192  & 7 & 7& 16 & 64 & 32 \\
         Switch Transformer  & 2.0  & 9.03  & 2048  & 8192  & 7& 7 & 16 & 32 & 32\\
         Base Layers & 2.0  & 10.31  &  2048  & 8192  & 8  & 8 & 16 & 32 &32\\
         HASH Layers  & 2.0  & 10.29  &  2048  & 8192  & 8  & 8 & 16 & 32 & 32 \\
         \midrule
         Switch Transformer - Enlarge  & 2.3  & 10.31  & 2048  & 8192  & 8  & 8 & 16 & 32 & 32 \\
         \midrule
         sMLP -- deterministic (our model) & 2.0 & 9.41  & 2048  & 8192  & 7  & 22 & 1 & 32 & 32\\
         \bottomrule
    \end{tabular}}
    \caption{\textbf{Scaled-up Model details}: We control FLOPs to the same level by adjusting the number of layers of the model. Since our model is based on gMLP \citep{gmlp}, more dense layers are needed, and only one head is needed for each device. While controlling FLOPs value, we found that only the Switch Transformer \citep{switch} can be compared with our sMLP among all the baselines. With the same amount of FLOPs value, our model has more parameters than the Switch Transformer; then, we train a Switch Transformer - Enlarge model for comparison.}
    % \vspace{-2mm}
    \label{tab:bigger_size_model_settings}
\end{table*}

\vspace{-2mm}
\paragraph{Implementation}
Our sMLP model contains $N_1$ dense blocks and $N_2$ sparse blocks. We follow Base Layers \citep{baselayer} setting: insert sparse blocks after the $\lfloor\frac{(N_1 + N_2)}{N_2+1}\rfloor\dots\lfloor\frac{N_2 (N_1 + N_2)}{N_2+1}\rfloor$th dense layers. For sparse blocks, it contains one tMoE module and one sMoE module. We adopt tMoE modules from Base Layers \citep{baselayer}. For our sMoE module, in each expert module, we insert one Spatial Gating Unit. Among our baselines, Base Layers \citep{baselayer} and HASH Layers \citep{hashlayer} adopt the same approach to arranging sparse layers. In contrast, Gshard \citep{gshard} and Switch Transformer \citep{switch} insert a sparse module at every other layer. More details of model architectures and implementation details can be found in Appendix~\ref{sec:more_experiments}. 

% \vspace{-2mm}

% \mikel{It's good to have extra info in the appendix, but you should give more details about the model sizes here. I would have a table with all the models you compare and some basic info (nlayers, hiddendim, param count etc) in the main paper.}

% \subsection{Expressiveness of MLP}

\vspace{-2mm}
\subsection{Token-wise operations comparison}
\label{sec:dense_model_comparision}

In this section, we compare our model with two dense models: Transformer \citep{vaswani2017attention} and gMLP \citep{gmlp}. The main difference between the all-MLP-based and the Transformer-based models is the token-wise operation. We compare these three kinds of token-wise operations: the self-attention module in Transformers, the Spatial Gating Unit in gMLP, and the sMoE module in our model.

In Table~\ref{tab:token_wise}, we compare these three token-wise operations and their respective head mechanisms. To extend multi-head mechanism into gMLP mode, instead of using one $W_s$ in Equation (\ref{eq:gmlp}), we chunk the input $X$ into $h$ parts along hidden dimension, and each part is multiplied by a separate $W_s$. We set head number $h$ to be 16 to compare Spatial Gating Unit from gMLP with Self-attention module from Transformer. Then the total number of parameters is $h*T*T$. Compared to Transformer self attention, a shortcoming of the Spatial Gating Unit is that the number of parameters increases linearly with the number of heads $h$ and quadratically with sequence length $T$. 
For comparison, our s-MoE module set one expert on each device, and each expert only contains one $W_s$, which is equal to heads number 1.

In Fig.~\ref{fig:dense_comparison}, we compare our model with the dense model with different heads number. The Transformer model greatly benefits from the multi-head mechanism. However, although the gMLP model increases the parameter amount, it does not improve performance through the multi-head mechanism. Our model can also be seen as a solution for gMLP's multi-head. Our model dramatically improves the performance of the all-MLP-based model and also outperforms the Transformer model.

\vspace{-2mm}
\subsection{Results on Sparse MLP}
\label{sec:small_model_results}

We report quality (valid perplexity) and training efficiency in Fig.~\ref{fig:model_comparision}, measured by number of training steps (\textbf{Top}) and training time (\textbf{Bottom}). We found that sMLP with both variants of routing strategies outperforms state-of-the-art Transformer-based MoE models with roughly the same amount of FLOPs. We point out that sample efficiency observed on a step basis does not necessarily translate into better model quality as measured by the wall clock due to two reasons. 
First, our model has additional all2all communication cost for sMoE module. Second, Base Layers, HASH Layers, and our model send tokens/vectors to the expert module in a balanced manner. In contrast, Gshard and Switch Transformer leverage a loss function to balance the assignment, which is not an equally balanced assignment. Although they have the same FLOPs, unbalanced loading can add extra computing time.
% communication costs. \mikel{unbalanced loading and communication cost are unrelated concepts, even if communication was instantaneous unbalanced loading could still be a bottleneck} %Therefore, the sample efficiency observed on a step basis does not necessarily translate to better model quality as measured by the wall clock. 
Fig.~\ref{fig:model_comparision} (\textbf{Bottom}) shows that given the same training time, our model achieves the best results (lowest valid perplexity), indicating that our sparse all-MLP models improved training efficiency over state-of-the-art Transformer-based MoE models on language modeling. %that our model performs better than all the other sparse models for fixed training duration and computational budget. 

Table~\ref{tab:pre-training-results} (\textbf{Top}) summarizes the detailed comparison in the main experiments. We control the FLOPs of all models to be roughly 0.8T. Except that the number of model layers is different, their embedding dimension is 1024, and the hidden dimension is 4096. We can see that our model achieves the best generalization at 25k training steps and in the meanwhile achieves the highest training speed. HASH Layers has the best performance among all Transformer baselines and requires the least time, which is consistent with results reported in \citet{hashlayer}.

% Except our model, all other models have 16 heads.

% \mikel{These are the main results in the paper and I think they deserve more discussion. In particular, you should also say something about how the different baselines do and whether this is consistent with what we would expect from the literature.}

% \subsection{Rocketing to large scale}
\subsection{Evaluation on Scalability}
\label{sec:large_scale}

% \begin{figure}
%     \centering
%     \includegraphics[width=0.7\linewidth]{iclr2022/img/large_model_fig.png}
%     \caption{Large Model Comparison}
%     \label{fig:large_model_comparision_fig}
% \end{figure}

In order to test the scalability of our model, we increase the model size training for 2.0 TFLOPs. Table~\ref{tab:pre-training-results} (\textbf{Bottom}) summarizes the results.

Compared to models in Table~\ref{tab:pre-training-results} (\textbf{Top}), we enlarged all of the models, changing the embedding from 1024 to 2048, and adjusting the hidden dimension from 4096 to 8192 as reported in Table~\ref{tab:bigger_size_model_settings}. We also increased the pretraining data size as described in Table~\ref{tab:datasets}.
% We also increased the pretraining data by a factor of 10. 
% \mikel{experimental settings say you are using 10x roberta + cc100en, so that should be more than 10x. also there is a lot of repetition throughout the paper, eg no need to repeat things that you already discussed in the experimental settings here, you can just refer to that} %\xian{do you still have the number of tokens for small scale experiments?} \ping{The large scale dataset is divided to be 10 parts. And the small dataset is one part, which contains 1/10 data.}

% When controlling FLOPs value to be 2.0T, our model outperforms all baselines in terms of perplexity. 
In this setting, we found Switch Transformer to remain a strong baseline. Although Base Layers and HASH Layers perform well when trained on small models and datasets, we find stability issues on large-scale training. Although previous papers \citep{switch,gshard} have controlled the FLOPs for model comparison, we found the FLOPs-match Switch Transformer has fewer parameters than our model. Therefore, to provide a more fair comparison, we also trained an enlarged version of Switch Transformer (by adding one additional sparse layer and dense layer) with 10.31B parameters which also increases its FLOPs to 2.3T. Our sMLP still outperforms Switch Transformer-Enlarge despite the latter having more FLOPs. %The enlarged Switch Transformer model will take longer to train, and the speed will be slower. Nevertheless, the quality (perplexity) is not as good as our model. 

% \begin{savenotes}
\begin{table*}[t!]
\centering
\scalebox{0.82}{
\begin{tabular}{l| c c c c c c c}
    \toprule
        Models & COPA &  PIQA & StoryCloze & Winogrande & HellaSwag & ReCoRD & Average\\
        \midrule
        \textcolor{dgray}{GPT-3 \citep{gpt3} from paper} & \textcolor{dgray}{73.0} & \textcolor{dgray}{72.9} & \textcolor{dgray}{72.4} & \textcolor{dgray}{57.4} & \textcolor{dgray}{51.0} & \textcolor{dgray}{82.1} & \textcolor{dgray}{68.13} \\
        \midrule
        Gshard \citep{gshard}& 76.00 & 68.12 & 67.88 & 51.068 & 38.00 & 72.39 & 62.24\\
        Switch Transformer \citep{switch} & 75.0 & 72.96 & 73.33 & 53.43& 52.48 & \textcolor{tblue}{\textbf{79.86}} &67.84 \\
        Base Layers \citep{baselayer} & 63.00 & 63.82 & 61.41 & 50.98 & 30.22 & 60.70 &55.02\\
        HASH Layers \citep{hashlayer} & 64.00 & 63.77 & 64.72 & 51.70 & 33.04 & 67.15 &57.40\\
        \midrule
         sMLP -- deterministic (our) & \textcolor{tblue}{\textbf{79.00}} & \textcolor{tblue}{\textbf{72.96}} & \textcolor{tblue}{\textbf{74.67}} & \textcolor{tblue}{\textbf{54.31}} & \textcolor{tblue}{\textbf{54.53}} & 73.42& \textcolor{tblue}{\textbf{68.15}} \\
         \bottomrule
    \end{tabular}}
    \caption{\textbf{Zero-shot priming evaluation}: we provide head-to-head comparison of our sMLP model with FLOPs-matched state-of-the-art sparse Transformers on six representative NLP tasks evaluated in GPT-3 in-context learning\citep{gpt3}. Specifically, we control for training data, batch size, learning rate and training steps for all these models. As a reference point, we also compare to the performance of a FLOPs-matched GPT3 model (dense Transformer) reported in \citet{gpt3}. Note that the GPT3 model from the paper uses larger training dataset (about 3 $\times$ larger).}\label{tab:zero-shot-evaluation}
    % \vspace{-2mm}
\end{table*}
% \end{savenotes}

\subsection{Zero-shot Priming}
\label{sec:zero_shot}
The previous experiments demonstrate sMLP's strong performance in language modeling. In this section,  we evaluate whether these gains in pretraining translate to improved generalization in downstream tasks. Specifically, we measure zero-shot in-context learning capability as is shown in GPT-3 \citep{gpt3}.

\vspace{-2mm}
\paragraph{Baselines}
Our baselines are Gshard \citep{gshard}, Switch Transformer \citep{switch}, Base Layers \citep{baselayer} and HASH Layers \citep{hashlayer} with roughly 2.0 TFLOPs from Section~\ref{sec:large_scale}.

\vspace{-2mm}
\paragraph{Tasks and datasets}
We select six representative NLP tasks which probes commonsense reasoning and question answering, including COPA \citep{roemmele2011choice}, PIQA \citep{bisk2020piqa}, StoryCloze \citep{mostafazadeh2016corpus}, Winogrande \citep{levesque2012winograd}, HellaSwag \citep{zellers2019hellaswag}, ReCoRD \citep{zhang2018record}. 
We use a language model to separately score each label choice using the same templates, and pick the one with the
highest score as \citet{artetxe2021efficient}.
More details about tasks and datasets can be found in Appendix~\ref{sec:app_downstream_task_description}. 

% \mikel{you should say something about the scoring function, templates etc. you can refer to an existing paper if you follow their setup (most likely our 1.1t paper?)}

\vspace{-2mm}
\paragraph{Results}
As is shown in Table~\ref{tab:zero-shot-evaluation}, sMLP outperforms all sparse Transformers in terms of average accuracy. Notable improvements come from commonsense reasoning tasks such as COPA, StoryCloze and HellaSwag. % and achieves state-of-the-art performance on most tasks except for ReCoRD task.
% Our model outperforms baselines on the PIQA task, StoryCloze task, and Winogrande task. Our model lags the Switch Transformer model \citep{switch} on the ReCoRD task.
We also compared to a FLOPs-matched dense Transformer reported in the GPT-3 paper \citep{gpt3}, which served as the strongest dense baseline. It is worth noting that the GPT-3 model was trained with more pre-training data (GPT-3 used 300 billion tokens for pre-training and our pre-training data contains 100 billion tokens).

%% file: content/6_related_works.tex
\section{Related Work}
\label{sec:related_works}

Mixture of Experts (MoE) was shown to be effective in \citet{shazeer2017outrageously}, where an MoE layer was stacked between LSTM \citep{hochreiter1997long} layers. More recently, Gshard \citep{gshard} extended this idea to Transformer \citep{vaswani2017attention} feed-forward layers and provided a way for parallel computing, which scaled up multilingual neural machine translation with Sparsely-Gated Mixture-of-Experts beyond 600 billion parameters using automatic parameter sharding.

% \xian{Add more specifics, e.g. data sharding or parameter sharding?}. 

Several recent work improves the routing strategy in Transformer-based MoEs. Switch Transformer \citep{switch} shows the design can be simplified by routing tokens to only one single expert (top-1 routing). 
% The switch transformer adopts a selective precision method, smaller parameter initialization, and high expert dropout to overcome the instability problem. 
In addition, they design a differentiable load balancing loss to reduce communication latency of sending tokens to different devices. Base Layers \citep{baselayer} further simplifies the framework by eliminating the balancing function, adapting the balancing assignment algorithm \citep{bertsekas1992auction} to send tokens from each batch to different devices equally. Instead of learning the gating weights for token assignment, HASH Layers \citep{hashlayer} design a Hash function to replace the learnable gating weights.
\cite{jaszczur2021sparse} is the first one successfully extending sparsity to attention layers. However, their results could just match dense transformer and did not get improvements in training speed.
Different from all of the above routing methods, our sMoE-deterministic method (described in Section~\ref{sec:deterministic routing}) chunk vectors by deterministic assignment.

% \mikel{maybe establish a link with our deterministic routing?}

There is also a growing body of work studying the overall design and scaling of Transformer-based MoE models. \citet{alibaba} explores key factors inside sparse expert models and investigates how they affect the model quality and computational efficiency. \citet{clark2022unified} studied the scaling laws of MoE models in terms of effective parameter count and compared three routing strategies. \citet{zoph2022designing} investigated several training details which affects training stability and finetuning performance. In addition to NLP, \citet{riquelme2021scaling} firstly applied MoEs in the Computer Vision field.

%The aforementioned models are all exploring how to perform model effective gating and token assignments, which is based on the transformer \citep{vaswani2017attention} structure, and is pre-defined for the FFN module. 
A closely related work is \citet{lou2021sparse}, which developed all-MLP-based MoEs for computer vision. When applied to NLP, there are additional challenges in designing the sparsely activated token-mixing operations by preserving the sequential nature of the input, which would otherwise lead to information leaking as is shown in Section \ref{subsec:routing_challenges}.
%, and our model is designed for the NLP field. Although transformer-based models are developing in the computer vision field, CNN is still dominant. We could think “MLP-Mixer” \citep{mlp_mixer} equal to “conv layer with 1*1 kernels”. Therefore, MLP architectures in the computer vision field are not very difficult. However, in the natural language field, self attention is much more important. We compared the gMLP \citep{gmlp} with dense Transformer \citep{vaswani2017attention} with the same amount of model parameters. The performance of gMLP \citep{gmlp} is not comparable with transformer. It will be more difficult to achieve good results with a complete MLP-based MoE in the NLP field. In addition, in the NLP field, we need to pay much attention to token sequence, and we cannot let the token leverage future token information (as we discussed in Sec.\ref{sec:sparse_mlp}. More details can be found in Appendix \ref{sec:challenges}), which is a great challenge. 

% In order to solve this special problem, we proposed two additional solutions: deterministic routing (Sec.\ref{sec:deterministic routing}) and partial prediction (Sec.\ref{sec:partial_prediction}). In contrast, it does not have this concern, and the current MoE method can be directly used on above of MLP \citep{mlp_mixer} structure.

%% file: content/7_conclusion.tex
\section{Conclusion}
\label{sec:conclusion}

In this work, we proposed sMLP, extending the recent gMLP \citep{gmlp} model with sparsely activated conditional computation using mixture-of-experts (MoEs). Different from Transformer-based MoEs, which only contains sparse feed-forward layers while still keeping dense computation for self-attention, the proposed architecture is fully sparse. We analyzed the challenges in designing routing strategies of sparse all-MLP architecture for language modeling and proposed two solutions. Through extensive evaluations on language modeling, we show that sMLP outperforms state-of-the-art sparse Transformer-based MoE models in terms of generalization and 2$\times$ improvement in training efficiency. Besides gains in pretraining, sMLP also achieves higher accuracy in zero-shot in-context learning of six representative NLP tasks, closing the gap of all-MLP architecture with Transformers as was observed in gMLP.   
It is worth noting that all our discussions about routing challenges and solutions are based on autoregressive language modeling. In the future, we hope to try all-MLP based MoE method with encoder-only models with bidirectional attention.

%% file: content/A_other_methods.tex
\section{Direct applying token-wise routing to gMLP}
\label{sec:challenges}

\begin{figure}[ht!]
    \centering
    \includegraphics[width=\linewidth]{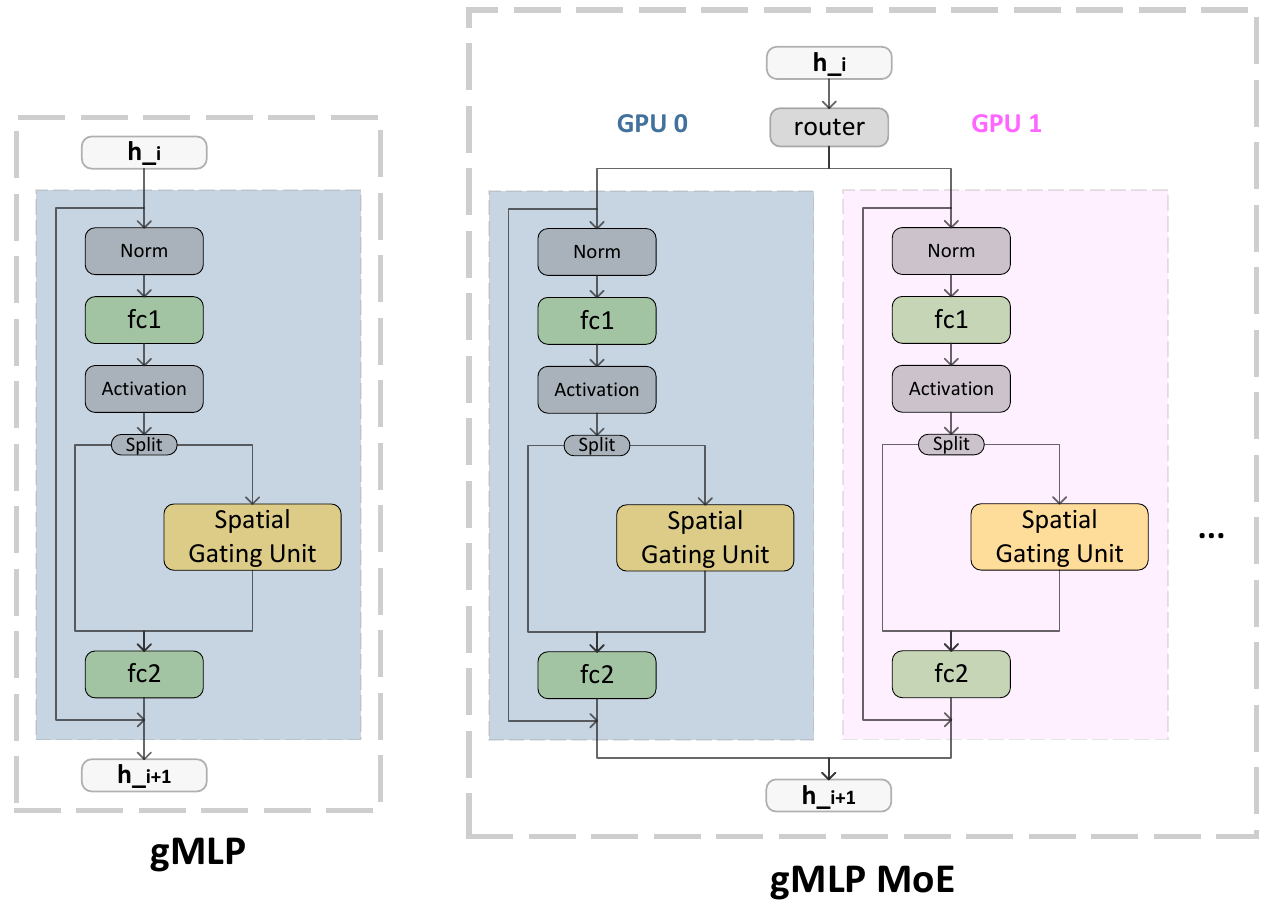}
    \vspace{-5mm}
    \caption{\textbf{Left}: one layer of gMLP\citep{gmlp}; \textbf{Right}: directly extend gMLP to token-level conditional computation commonly used in Transformer MoEs \citep{gshard,switch,baselayer,hashlayer} with two experts (GPUs).}
    \label{fig:gmlp_moe}
\end{figure}

%We discuss the drawbacks of directly applying  token-wise routing to all-MLP architecture such as gMLP. 
% Our goal is to extend an fully MLP architecture with mixture of experts. 
The most straightforward way to extend gMLP \citep{gmlp} to mixture-of-experts (MoEs) is to apply the same token-routing used in Transformer-based MoEs \citep{gshard,switch,baselayer,hashlayer}, as is illustrated in Fig.~\ref{fig:gmlp_moe} in the case of two experts (GPUs). % The gMLP \citep{gmlp} is a dense model, in that way the model is the same among GPUs and there are no interactions between GPUs during forward computation, as shown in the left of Fig.\ref{fig:gmlp_moe}. The gMLP MoE leverages the standard MoE way to send each tokens to different experts according to router. The right figure in Fig.\ref{fig:gmlp_moe} shows the example with 2 GPUs. Each expert module is a Spatial Gating Unit. 

% \xian{I still feel it's tricky to position this section clearly w/o distracting main message.}
% \begin{wraptable}{l}{6cm}
% \begin{tabular}{l| c c}
%     \toprule
%         Models & GPUs & Valid Perplexity \\
%         \midrule
%          gMLP & 32 & 15.7 \\
%          gMLP MoE & 32 & 17.2 \\
%          gMLP MoE & 64 & 20.5 \\
%          \bottomrule
%     \end{tabular}
%     \caption{gMLP MoE.}\label{tab:gmlp_moe}
% \end{wraptable}

\begin{table}[ht!]
\centering
\scalebox{0.8}{
\begin{tabular}{l| c c}
    \toprule
        Models & \# Experts & Valid Perplexity \\
        \midrule
         gMLP & 0 & 15.7 \\
         gMLP MoE & 32 & 17.2 \\
         gMLP MoE & 64 & 20.5 \\
         \bottomrule
    \end{tabular}}
    \caption{Extending gMLP to token-based MoEs hurts performance compared to dense gMLP. Furthermore, increasing the number of experts worsens the performance.}\label{tab:gmlp_moe}
\end{table}

Table~\ref{tab:gmlp_moe} shows that this approach, despite being simple and straightforward, has poor generalization. Since the Spatial Gating Unit is inserted into each device, and the tokens from the same sentence will be sent to different devices. That means tokens can only attend to previous tokens which were sent to the same device. This problem will be more severe as we increase expert (device) numbers. As the number of experts increases, the probability of sending tokens to the same expert from one sentence will decrease, so the number of tokens that each word can attend will decrease. 

%Therefore, with the increase of GPU devices, the performance will be worse. This is not the effect we want to see.

% \begin{table*}[t!]
% \centering
% \scalebox{0.8}{
% \begin{tabular}{l| c c c c c c c c}
%     \toprule
%         Models  & \makecell{Model Size \\ (Parameters)}&\makecell{Quality after 25k steps \\(Perplexity $\downarrow$)} &  \makecell{Time to Quality\\ (hours $\downarrow$)} & \makecell{Speed\\ (world per second $\uparrow$)} \\
%         \midrule
%         Gshard \citep{gshard} & 3.48B & 19.26 & 27.54 & 136k \\
%         Switch Transformer \citep{switch} & 3.48B & 15.31 & 20.67 & 185k \\
%         Base Layers \citep{baselayer} & 3.60B &18.16 & 21.13 & 178k \\
%         HASH Layers \citep{hashlayer} & 3.51B &13.57 & 21.08 & 176k \\
%         \midrule
%         sMLP & 3.50B & 8.01 & 21.45 & 170k \\ 
%         sMLP -- deterministic & 3.50B & 13.46 & 20.41 & 192k\\
%         sMLP -- partial & 3.50B & 14.26 & 20.45 & 186k \\
%          \bottomrule
%     \end{tabular}}
%     \caption{Compared sMLP method with sMLP -- deterministic (Sec.\ref{sec:deterministic routing}) and sMLP -- partial (Sec.\ref{sec:partial_prediction}). \ping{change to different figure.}}
%     \label{tab:smlp}
% \end{table*}

\begin{figure*}[ht!]
    \centering
    \includegraphics[width=0.8\linewidth]{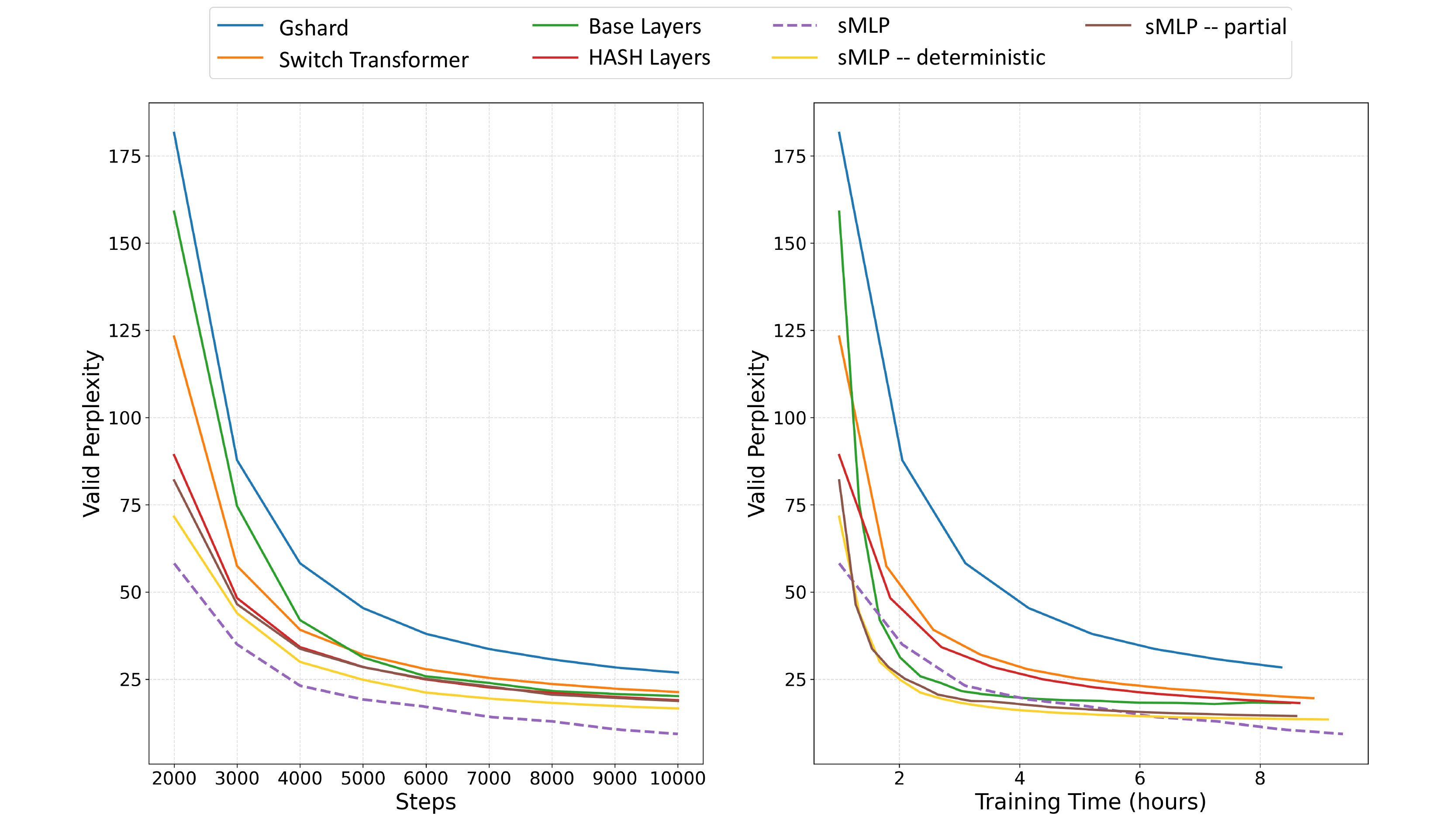}
    \caption{Compared sMLP method with sMLP -- deterministic (Section~\ref{sec:deterministic routing}) and sMLP -- partial (Section~\ref{sec:partial_prediction})}
    \label{fig:smlp}
\end{figure*}

\section{Gating analysis for the sMLP module}
\label{sec:normal_gating}
% \mikel{Again I would completely remove everything from this point forward until sec 3.2.1: the approach you are describing is not theoretically sound and so it is very confusing for the reader. You can replace this with a short paragraph that explains that it is challenging to route hidden vectors while making sure that we don't use information from future tokens, and say that we explore two solutions: deterministic routing and partial prediction.}

%We first try to use the previous MoE gating method to learn how to allocate the hidden vector. 
MoE layers take token representation $X \in \mathbb{R}^{T \times H}$, where $T$ represents the sequence length and $H$ represents the hidden dimension. In the sMoE, we transpose input to the hidden vector $V \in \mathbb{R}^{H*T}$. A gating function routes this to the best top-$k$ experts, selected from a set $\{E_i(v)\}_{i=1}^N$ of $N$ experts. The router variable $W_r \in \mathbb{R}^{T \times N}$ produces $h(V) = V W_r$ which are normalized via a softmax distribution over the available $N$ experts at that layer. $V$ contains $H$ hidden vector $v_i\in \mathbb{R}^{T}$. The probability for sending $i_{th}$ hidden vector is given by
\begin{equation}
    p_i(v) = \frac{e^{h(v)_i}}{\sum_j^N e^{h(v)_j}}
\label{eq:smoe_gating}
\end{equation}
The top-$k$ gate values are selected for routing the hidden vector $v$. If $\mathcal{T}$ is the set of selected top-$k$ indices then the output computation of the layer is the linearly weighted combination of each expert's computation on the token by the gate value,
\begin{equation}
    y = \sum_{i \in \mathcal{N}} p_i(v) E_i(v) 
\end{equation}

Through learning the gating weight $W_r$, the MoE model routes hidden vectors to different experts. 

Fig.~\ref{fig:smlp} shows the results by this method (sMLP in the figure). We can see that this method achieves the lowest perplexity. However, this is due to information leaking from future tokens in learning $W_r$. Specifically, the gating weight for sMoE is $W_r \in \mathbb{R}^{T\times N}$, which is trained after seeing all the tokens in the sentence, especially looking ahead at future tokens. In this way, this gating can making "smartest" decision.
This results in information leaking for autoregressive language models. Unlike self attention model, where a diagonal mask could be applied to attention scores $A \in \mathbb{R}^{T\times T}$  to prevent information leaking from future tokens, gating weights $W_r \in \mathbb{R}^{T\times N}$ cannot use a valid mask. Due to the aforementioned problems, we cannot adopt existing MoE gating methods for routing hidden vectors in autoregressive language models. Therefore, we propose two improved methods sMLP -- deterministic (in Section~\ref{sec:deterministic routing}) and sMLP --partial (in Section~\ref{sec:partial_prediction}).

%We have adopted the following two solutions: deterministic routing and partial prediction.

\section{More implementation details}
\label{sec:more_experiments}

\subsection{Choice of baselines}
\label{sec:app_baselines}
The dense counterpart is gMLP \citep{gmlp}. %A sparse block contains a tMoE module from BASE Layers \citep{baselayer}. 
We adapted the balanced assignment routing method proposed in Base Layers \citep{baselayer} for the tMoE implementation. Thus, the closest Transformer-based sparse models is Base Layers \citep{baselayer}. We also compared to the dense Transformer \citep{vaswani2017attention} and other state-of-the-art Transformer MoE models (Gshard \citep{gshard}, Base Layers \citep{baselayer}, HASH Layers \citep{hashlayer}) as listed in Table~\ref{tab:baselines}.

% \begin{table*}[ht!]
%     \centering
%     \scalebox{1.0}{
%     \begin{tabular}{l|c c c}
%     \toprule
%         Models & Speed (word per second $\uparrow$) & Quality (Valid Perplexity $\downarrow$) \\
%         \midrule
%          Gshard \citep{gshard}& 157 k &20.05 \\ 
%          Switch Transformer \citep{switch}& 171 k &14.93 \\ 
%          Base Layers \citep{baselayer}& 211 k &17.9 \\ 
%          Hash Layers \citep{hashlayer}& 458 k &15.3 \\
%          \midrule
%          sMLP &117 k& \textbf{8.14}\\ 
%          sMLP-deterministic &467 k&13.24 \\ 
%          sMLP-partial &\textbf{626 k}&14.45 \\
%          \bottomrule
    
%     \end{tabular}}
%     \caption{Model comparison.}
%     \label{tab:small_models}
% \end{table*}

% \begin{table*}[ht!]
%     \centering
%     \scalebox{1.0}{
%     \begin{tabular}{l|c c c}
%     \toprule
%         Models & Speed (word per second $\uparrow$) & Quality (Valid Perplexity $\downarrow$) \\
%         \midrule
%          Gshard \citep{gshard}& 223 k & 14.5 \\ 
%          Switch Transformer \citep{switch}& 219 k & 10.11 \\ 
%          Base Layers \citep{baselayer}& 236 k & 20.6 (divergence) \\ 
%          HASH Layers \citep{hashlayer}& 385 k & 16.69 \\
%          \midrule
%          sMLP-deterministic & 361 k & 9.06 \\ 
%          \bottomrule
%     \end{tabular}}
%     \caption{Large model comparison.   }
%     \label{tab:large_model}
% \end{table*}
\subsection{Training}
\paragraph{Small Model Configurations} Table~\ref{tab:small_model_architecture} shows the the detailed model architecture used in the main experiments. Through observation, we noticed that all the models (except for Gshard) drop negligibly after 15k. In addition, there is a small amount of fluctuation for the Base Layers after 25k. Then we choose 25k for model comparison in Table~\ref{tab:pre-training-results}.

\begin{table}[ht!]
    \centering
    \scalebox{0.75}{
    \begin{tabular}{l|c c c}
    \toprule
        Models & Base Layers & sMLP-deterministic \\
        \midrule
        Expert Parameters & 100.75 M & 113.44 M \\
        Non Expert Parameters & 241.79 M & 258.73 M \\
        FLOPs & 0.83 T & 0.83 T \\
        \midrule
        Expert Modules per MoE layer & 32 &32 \\
        Number of MoE layers & 12 & 12 \\
        FFNs per Expert Module & 1 & 1 \\
        Spatial Gating Unit per Expert Module & 1 & 1 \\
        Embedding Size & 1024 & 1024 \\
        Hidden Dimension & 4096 & 4096 \\
        Heads & 16 & 1 \\
        Number of Dense Layers & 12 & 28 \\
         \midrule
         Context Length & 1024 & 1024 \\
         Batch Size & 4 & 4 \\
         Gradient Accumulation & 4 & 4 \\
         \midrule
         Dropout & 0.1 & 0.1 \\
         ddp-backend & no\_c10d & no\_c10d \\
         optimizer & adam & adam\\
         adam-betas & (0.9, 0.98) & (0.9, 0.98) \\
         weight-decay & 0.1 & 0.1 \\
         lr & 5e-4 & 5e-4 \\
         lr-scheduler & inverse\_sqrt & inverse\_sqrt\\
         warmup-updates & 4000 & 4000 \\
         warmup-init-lr & 1e-7 & 1e-7 \\
         \bottomrule
    \end{tabular}}
    \caption{Model architectures used in the main experiments in Section~\ref{sec:small_model_results}.}
    \label{tab:small_model_architecture}
\end{table}

\begin{table}[ht!]
    \centering
    \scalebox{0.75}{
    \begin{tabular}{l|c c c}
    \toprule
        Models & Transformer & gMLP & sMLP-deterministic\\
        \midrule
        Expert Parameters & 0 M & 0 M &  258.73M\\
        Non Expert Parameters & 354.76 M & 353.87 M &113.44M \\
        FLOPs & 0.83 T & 0.83 T & 0.83 T \\
        \midrule
        Sparse Layers & 0 & 0 & 12 \\
        Dense Layers & 24 & 41 & 28 \\
        Embedding Size & 1024 & 1024 & 1024\\
        Hidden Dimension & 4096 & 4096 & 4096\\
        Heads & 16 & 1 & 1\\
         \midrule
         Context Length & 1024 & 1024 & 1024\\
         Batch Size & 2 & 2 & 2\\
         Gradient Accumulation & 4 & 4 & 4\\
         \midrule
         Dropout & 0.1 & 0.1 & 0.1\\
         ddp-backend & no\_c10d & no\_c10d & no\_c10d\\
         optimizer & adam & adam\\
         adam-betas & (0.9, 0.98) & (0.9, 0.98) & (0.9, 0.98)\\
         weight-decay & 0.1 & 0.1 & 0.1\\
         lr & 5e-4 & 5e-4& 5e-4 \\
         lr-scheduler & inverse\_sqrt & inverse\_sqrt & inverse\_sqrt \\
         warmup-updates & 4000 & 4000  & 4000\\
         warmup-init-lr & 1e-7 & 1e-7 & 1e-7\\
         \bottomrule
    
    \end{tabular}}
    \caption{Small dense model architectures used in Section~\ref{sec:dense_model_comparision}.}
    \label{tab:small_dense_model_architecture}
\end{table}

\paragraph{Large model configuration} Table~\ref{tab:large_model_architecture} shows the the model architecture used in scalability experiments. Through observation, we noticed that all the models (except for Base Layers) are converged around 70k. After 70k, these models drop negligibly. Then we choose 100k for model comparison in Table~\ref{tab:pre-training-results}.

\begin{table}[ht!]
    \centering
    \scalebox{0.75}{
    \begin{tabular}{l|c c c}
    \toprule
        Models & Base Layers & sMLP-deterministic \\
        \midrule
        Expert Parameters & 642.6 M & 682.6 M \\
        Non Expert Parameters & 268.6 M & 277.1 M \\
        FLOPs & 2.125 T & 2.125 T \\
        \midrule
        Expert Modules per MoE layer & 32 &32 \\
        Number of MoE layers & 8 & 8 \\
        FFNs per Expert Module & 1 & 1 \\
        Spatial Gating Unit per Expert Module & 1 & 1 \\
        Embedding Size & 2048 & 2048 \\
        Hidden Dimension & 8192 & 8192 \\
        Heads & 16 & 1 \\
        Number of Dense Layers & 8 & 20 \\
         \midrule
         Context Length & 1024 & 1024 \\
         Batch Size & 2 & 2 \\
         Gradient Accumulation & 8 & 8 \\
         \midrule
         Dropout & 0.1 & 0.1 \\
         ddp-backend & no\_c10d & no\_c10d \\
         optimizer & adam & adam\\
         adam-betas & (0.9, 0.98) & (0.9, 0.98) \\
         weight-decay & 0.1 & 0.1 \\
         lr & 1e-3 & 1e-3 \\
         lr-scheduler & inverse\_sqrt & inverse\_sqrt\\
         warmup-updates & 4000 & 4000 \\
         warmup-init-lr & 1e-7 & 1e-7 \\
         \bottomrule
    
    \end{tabular}}
    \caption{Large model architectures used in the scalability experiments in Section~\ref{sec:large_scale} and Section~\ref{sec:zero_shot}.}
    \label{tab:large_model_architecture}
\end{table}

\paragraph{Dense model configuration} Table~\ref{tab:small_dense_model_architecture} shows the model architecture we use in Section~\ref{sec:dense_model_comparision}.

%\subsection{Hardware} 
All the models are trained on 32 Nvidia 32G V100 GPUs. %connected with Infiniband.% \footnote{Communication between workers may be different on other networking hardware}. 
\subsection{Datasets}
\label{sec:appendix_datasets}
The RoBERTa contains five English-language corpora of varying sizes and domains:
\begin{itemize}
\item BOOKCORPUS \citep{zhu2015aligning} plus English WIKIPEDIA. This is the original data used to train BERT. (16GB).
\item CC-NEWS, which collected from the English portion of the CommonCrawl News dataset \citep{ccnews}. The data contains 63 million English news articles crawled between
September 2016 and February 2019. (76GB after filtering)
\item OPENWEBTEXT \citep{openweb}, an open-source recreation of the WebText corpus described in \citep{radford2019language}. The text is web content extracted from URLs shared on Reddit with at least three upvotes. (38GB).
\item STORIES, a dataset introduced in \citep{trinh2018simple} containing a subset of CommonCrawl
data filtered to match the story-like style of
Winograd schemas. (31GB).
\end{itemize}

\section{Transformer vs. gMLP}
\begin{figure}[ht!]
    \centering
    \includegraphics[width=0.95\linewidth]{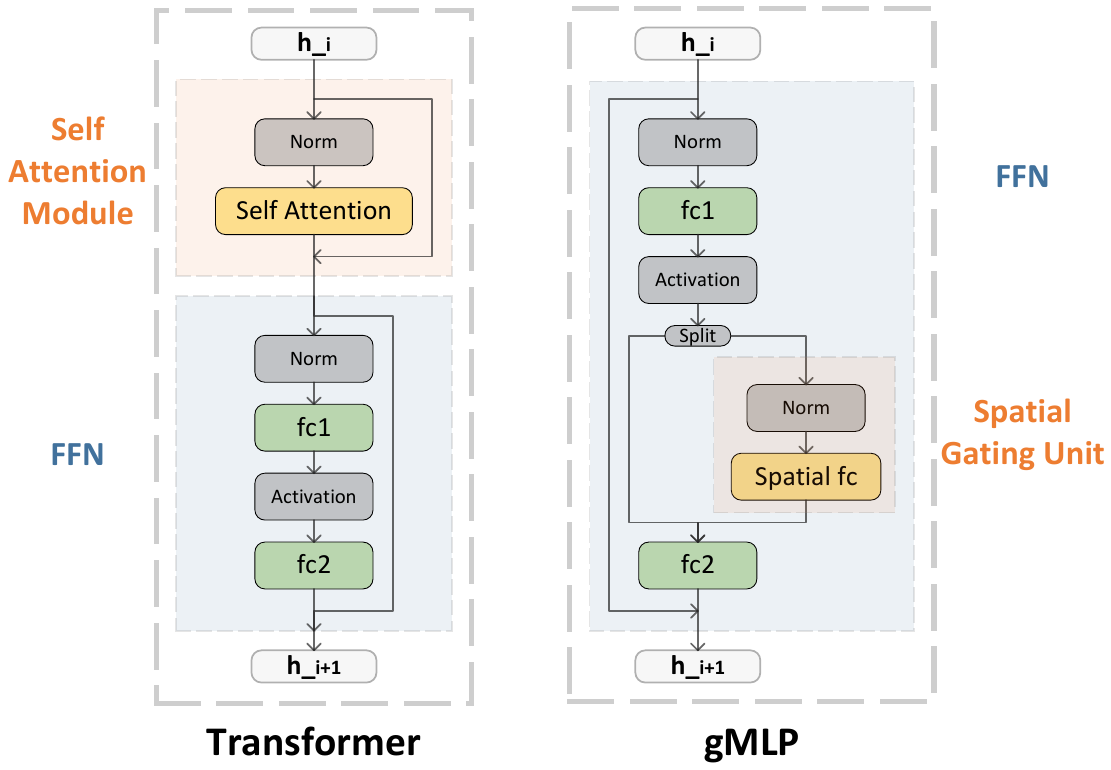}
    \vspace{-3mm}
    \caption{Model architecture comparison: Transformer v.s. gMLP}
    \label{fig:transformer_gmlp}
\end{figure}

We compared the Transformer \citep{transformer2017} with the gMLP \citep{gmlp} model architecture in Fig.~\ref{fig:transformer_gmlp}. When comparing these two model structures, the FFN module is the same. The differences exist in the token-wise operations. The Transformer leverages the self attention module to calculate the relations between different tokens. However, the gMLP model leverages the Spatial Gating Unit to replace the self attention module.

% \xian{This section only has a figure. Please add some text.}

\section{Downstream Task Descriptions}
\label{sec:app_downstream_task_description}

% \paragraph{Tasks and datasets}
We evaluate on six representative tasks covering commonsense reasoning and question answering, including:
\begin{itemize}
    \item \textbf{COPA} (Choice of Plausible Alternatives) \citep{roemmele2011choice} is a causal reasoning task in which a system is given a premise sentence and must determine either the cause or effect of the premise from two possible choices. All examples are handcrafted and focus on blogs and photography-related encyclopedia topics.
    \item \textbf{PIQA} (PhysicalQA) \citep{bisk2020piqa} is a question answering task which asks common sense questions about how the physical world works and is intended as a probe of grounded understanding of the world. 
    \item \textbf{StoryCloze} \citep{mostafazadeh2016corpus} involves selecting the correct ending sentence for five-sentence long stories.
    \item \textbf{Winogrande} The Winograd Schemas Challenge \citep{levesque2012winograd} is a classical task in NLP that involves determining which word a pronoun refers to when the pronoun is grammatically ambiguous but semantically unambiguous to a human. Recently fine-tuned language models have achieved near-human performance on the original Winograd dataset, but more complex versions adversarially-mined Winogrande dataset \citep{sakaguchi2020winogrande} still significantly lag human performance. We test our performance on Winogrande.
    \item \textbf{HellaSwag} dataset \citep{zellers2019hellaswag} involves picking the best ending to a story or set of instructions. The examples were adversarially mined to be difficult for language models.
    \item \textbf{ReCoRD} (Reading Comprehension with Commonsense Reasoning Dataset) \citep{zhang2018record} is a multiple-choice QA task. Each example consists of a news article and a Cloze-style question about the article in which one entity is masked out. The system must predict the masked out entity from a list of possible entities in the provided passage, where the same entity may be expressed with multiple different surface forms, which are all considered correct. Articles are from CNN and Daily Mail. 
\end{itemize}

%% file: example_paper.bbl
\begin{thebibliography}{42}
\providecommand{\natexlab}[1]{#1}
\providecommand{\url}[1]{\texttt{#1}}
\expandafter\ifx\csname urlstyle\endcsname\relax
  \providecommand{\doi}[1]{doi: #1}\else
  \providecommand{\doi}{doi: \begingroup \urlstyle{rm}\Url}\fi

\bibitem[Artetxe et~al.(2021)Artetxe, Bhosale, Goyal, Mihaylov, Ott, Shleifer,
  Lin, Du, Iyer, Pasunuru, et~al.]{artetxe2021efficient}
Artetxe, M., Bhosale, S., Goyal, N., Mihaylov, T., Ott, M., Shleifer, S., Lin,
  X.~V., Du, J., Iyer, S., Pasunuru, R., et~al.
\newblock Efficient large scale language modeling with mixtures of experts.
\newblock \emph{arXiv preprint arXiv:2112.10684}, 2021.

\bibitem[Bertsekas(1992)]{bertsekas1992auction}
Bertsekas, D.~P.
\newblock Auction algorithms for network flow problems: A tutorial
  introduction.
\newblock \emph{Computational optimization and applications}, 1\penalty0
  (1):\penalty0 7--66, 1992.

\bibitem[Bisk et~al.(2020)Bisk, Zellers, Gao, Choi, et~al.]{bisk2020piqa}
Bisk, Y., Zellers, R., Gao, J., Choi, Y., et~al.
\newblock Piqa: Reasoning about physical commonsense in natural language.
\newblock In \emph{Proceedings of the AAAI Conference on Artificial
  Intelligence}, volume~34, pp.\  7432--7439, 2020.

\bibitem[Brown et~al.(2020{\natexlab{a}})Brown, Mann, Ryder, Subbiah, Kaplan,
  Dhariwal, Neelakantan, Shyam, Sastry, Askell, et~al.]{brown2020language}
Brown, T.~B., Mann, B., Ryder, N., Subbiah, M., Kaplan, J., Dhariwal, P.,
  Neelakantan, A., Shyam, P., Sastry, G., Askell, A., et~al.
\newblock Language models are few-shot learners.
\newblock \emph{arXiv preprint arXiv:2005.14165}, 2020{\natexlab{a}}.

\bibitem[Brown et~al.(2020{\natexlab{b}})Brown, Mann, Ryder, Subbiah, Kaplan,
  Dhariwal, Neelakantan, Shyam, Sastry, Askell, et~al.]{gpt3}
Brown, T.~B., Mann, B., Ryder, N., Subbiah, M., Kaplan, J., Dhariwal, P.,
  Neelakantan, A., Shyam, P., Sastry, G., Askell, A., et~al.
\newblock Language models are few-shot learners.
\newblock \emph{arXiv preprint arXiv:2005.14165}, 2020{\natexlab{b}}.

\bibitem[Clark et~al.(2022)Clark, Casas, Guy, Mensch, Paganini, Hoffmann,
  Damoc, Hechtman, Cai, Borgeaud, et~al.]{clark2022unified}
Clark, A., Casas, D. d.~l., Guy, A., Mensch, A., Paganini, M., Hoffmann, J.,
  Damoc, B., Hechtman, B., Cai, T., Borgeaud, S., et~al.
\newblock Unified scaling laws for routed language models.
\newblock \emph{arXiv preprint arXiv:2202.01169}, 2022.

\bibitem[Conneau et~al.(2019)Conneau, Khandelwal, Goyal, Chaudhary, Wenzek,
  Guzm{\'a}n, Grave, Ott, Zettlemoyer, and Stoyanov]{conneau2019unsupervised}
Conneau, A., Khandelwal, K., Goyal, N., Chaudhary, V., Wenzek, G., Guzm{\'a}n,
  F., Grave, E., Ott, M., Zettlemoyer, L., and Stoyanov, V.
\newblock Unsupervised cross-lingual representation learning at scale.
\newblock \emph{arXiv preprint arXiv:1911.02116}, 2019.

\bibitem[Devlin et~al.(2018)Devlin, Chang, Lee, and Toutanova]{devlin2018bert}
Devlin, J., Chang, M.-W., Lee, K., and Toutanova, K.
\newblock Bert: Pre-training of deep bidirectional transformers for language
  understanding.
\newblock \emph{arXiv preprint arXiv:1810.04805}, 2018.

\bibitem[Fedus et~al.(2021)Fedus, Zoph, and Shazeer]{switch}
Fedus, W., Zoph, B., and Shazeer, N.
\newblock Switch transformers: Scaling to trillion parameter models with simple
  and efficient sparsity.
\newblock \emph{arXiv preprint arXiv:2101.03961}, 2021.

\bibitem[Gokaslan \& Cohen(2019)Gokaslan and Cohen]{openweb}
Gokaslan, A. and Cohen, V.
\newblock Openwebtext corpus.
\newblock In \emph{http://web.archive.org/save/http://Skylion007.github.io/
  OpenWebTextCorpus.}, 2019.

\bibitem[Hochreiter \& Schmidhuber(1997)Hochreiter and
  Schmidhuber]{hochreiter1997long}
Hochreiter, S. and Schmidhuber, J.
\newblock Long short-term memory.
\newblock \emph{Neural computation}, 9\penalty0 (8):\penalty0 1735--1780, 1997.

\bibitem[Hou et~al.(2021)Hou, Jiang, Yuan, Cheng, Yan, and Feng]{hou2021vision}
Hou, Q., Jiang, Z., Yuan, L., Cheng, M.-M., Yan, S., and Feng, J.
\newblock Vision permutator: A permutable mlp-like architecture for visual
  recognition.
\newblock \emph{arXiv preprint arXiv:2106.12368}, 2021.

\bibitem[Jaszczur et~al.(2021)Jaszczur, Chowdhery, Mohiuddin, Kaiser, Gajewski,
  Michalewski, and Kanerva]{jaszczur2021sparse}
Jaszczur, S., Chowdhery, A., Mohiuddin, A., Kaiser, L., Gajewski, W.,
  Michalewski, H., and Kanerva, J.
\newblock Sparse is enough in scaling transformers.
\newblock \emph{Advances in Neural Information Processing Systems},
  34:\penalty0 9895--9907, 2021.

\bibitem[Lee-Thorp et~al.(2021)Lee-Thorp, Ainslie, Eckstein, and
  Ontanon]{lee2021fnet}
Lee-Thorp, J., Ainslie, J., Eckstein, I., and Ontanon, S.
\newblock Fnet: Mixing tokens with fourier transforms.
\newblock \emph{arXiv preprint arXiv:2105.03824}, 2021.

\bibitem[Lepikhin et~al.(2020)Lepikhin, Lee, Xu, Chen, Firat, Huang, Krikun,
  Shazeer, and Chen]{gshard}
Lepikhin, D., Lee, H., Xu, Y., Chen, D., Firat, O., Huang, Y., Krikun, M.,
  Shazeer, N., and Chen, Z.
\newblock Gshard: Scaling giant models with conditional computation and
  automatic sharding.
\newblock \emph{arXiv preprint arXiv:2006.16668}, 2020.

\bibitem[Levesque et~al.(2012)Levesque, Davis, and
  Morgenstern]{levesque2012winograd}
Levesque, H., Davis, E., and Morgenstern, L.
\newblock The winograd schema challenge.
\newblock In \emph{Thirteenth International Conference on the Principles of
  Knowledge Representation and Reasoning}, 2012.

\bibitem[Lewis et~al.(2021)Lewis, Bhosale, Dettmers, Goyal, and
  Zettlemoyer]{baselayer}
Lewis, M., Bhosale, S., Dettmers, T., Goyal, N., and Zettlemoyer, L.
\newblock Base layers: Simplifying training of large, sparse models.
\newblock \emph{arXiv preprint arXiv:2103.16716}, 2021.

\bibitem[Liu et~al.(2021{\natexlab{a}})Liu, Dai, So, and Le]{gmlp}
Liu, H., Dai, Z., So, D.~R., and Le, Q.~V.
\newblock Pay attention to mlps.
\newblock \emph{arXiv preprint arXiv:2105.08050}, 2021{\natexlab{a}}.

\bibitem[Liu et~al.(2021{\natexlab{b}})Liu, Dai, So, and Le]{liu2021pay}
Liu, H., Dai, Z., So, D.~R., and Le, Q.~V.
\newblock Pay attention to mlps.
\newblock \emph{arXiv preprint arXiv:2105.08050}, 2021{\natexlab{b}}.

\bibitem[Liu et~al.(2019)Liu, Ott, Goyal, Du, Joshi, Chen, Levy, Lewis,
  Zettlemoyer, and Stoyanov]{liu2019roberta}
Liu, Y., Ott, M., Goyal, N., Du, J., Joshi, M., Chen, D., Levy, O., Lewis, M.,
  Zettlemoyer, L., and Stoyanov, V.
\newblock Roberta: A robustly optimized bert pretraining approach.
\newblock \emph{arXiv preprint arXiv:1907.11692}, 2019.

\bibitem[Lou et~al.(2021)Lou, Xue, Zheng, and You]{lou2021sparse}
Lou, Y., Xue, F., Zheng, Z., and You, Y.
\newblock Sparse-mlp: A fully-mlp architecture with conditional computation.
\newblock \emph{arXiv preprint arXiv:2109.02008}, 2021.

\bibitem[Mostafazadeh et~al.(2016)Mostafazadeh, Chambers, He, Parikh, Batra,
  Vanderwende, Kohli, and Allen]{mostafazadeh2016corpus}
Mostafazadeh, N., Chambers, N., He, X., Parikh, D., Batra, D., Vanderwende, L.,
  Kohli, P., and Allen, J.
\newblock A corpus and evaluation framework for deeper understanding of
  commonsense stories.
\newblock \emph{arXiv preprint arXiv:1604.01696}, 2016.

\bibitem[Nagel(2016)]{ccnews}
Nagel, S.
\newblock Cc-news.
\newblock In \emph{http://web.archive.org/save/http://commoncrawl.org
  /2016/10/newsdataset-available.}, 2016.

\bibitem[Ott et~al.(2019)Ott, Edunov, Baevski, Fan, Gross, Ng, Grangier, and
  Auli]{fairseq}
Ott, M., Edunov, S., Baevski, A., Fan, A., Gross, S., Ng, N., Grangier, D., and
  Auli, M.
\newblock fairseq: A fast, extensible toolkit for sequence modeling.
\newblock \emph{arXiv preprint arXiv:1904.01038}, 2019.

\bibitem[Paszke et~al.(2017)Paszke, Gross, Chintala, Chanan, Yang, DeVito, Lin,
  Desmaison, Antiga, and Lerer]{pytorch}
Paszke, A., Gross, S., Chintala, S., Chanan, G., Yang, E., DeVito, Z., Lin, Z.,
  Desmaison, A., Antiga, L., and Lerer, A.
\newblock Automatic differentiation in pytorch.
\newblock 2017.

\bibitem[Radford et~al.(2018)Radford, Narasimhan, Salimans, and
  Sutskever]{radford2018improving}
Radford, A., Narasimhan, K., Salimans, T., and Sutskever, I.
\newblock Improving language understanding by generative pre-training.
\newblock 2018.

\bibitem[Radford et~al.(2019)Radford, Wu, Child, Luan, Amodei, Sutskever,
  et~al.]{radford2019language}
Radford, A., Wu, J., Child, R., Luan, D., Amodei, D., Sutskever, I., et~al.
\newblock Language models are unsupervised multitask learners.
\newblock \emph{OpenAI blog}, 1\penalty0 (8):\penalty0 9, 2019.

\bibitem[Raffel et~al.(2019)Raffel, Shazeer, Roberts, Lee, Narang, Matena,
  Zhou, Li, and Liu]{raffel2019exploring}
Raffel, C., Shazeer, N., Roberts, A., Lee, K., Narang, S., Matena, M., Zhou,
  Y., Li, W., and Liu, P.~J.
\newblock Exploring the limits of transfer learning with a unified text-to-text
  transformer.
\newblock \emph{arXiv preprint arXiv:1910.10683}, 2019.

\bibitem[Riquelme et~al.(2021)Riquelme, Puigcerver, Mustafa, Neumann, Jenatton,
  Pinto, Keysers, and Houlsby]{riquelme2021scaling}
Riquelme, C., Puigcerver, J., Mustafa, B., Neumann, M., Jenatton, R., Pinto,
  A.~S., Keysers, D., and Houlsby, N.
\newblock Scaling vision with sparse mixture of experts.
\newblock \emph{arXiv preprint arXiv:2106.05974}, 2021.

\bibitem[Roemmele et~al.(2011)Roemmele, Bejan, and Gordon]{roemmele2011choice}
Roemmele, M., Bejan, C.~A., and Gordon, A.~S.
\newblock Choice of plausible alternatives: An evaluation of commonsense causal
  reasoning.
\newblock In \emph{2011 AAAI Spring Symposium Series}, 2011.

\bibitem[Roller et~al.(2021)Roller, Sukhbaatar, Szlam, and Weston]{hashlayer}
Roller, S., Sukhbaatar, S., Szlam, A., and Weston, J.
\newblock Hash layers for large sparse models.
\newblock \emph{arXiv preprint arXiv:2106.04426}, 2021.

\bibitem[Sakaguchi et~al.(2020)Sakaguchi, Le~Bras, Bhagavatula, and
  Choi]{sakaguchi2020winogrande}
Sakaguchi, K., Le~Bras, R., Bhagavatula, C., and Choi, Y.
\newblock Winogrande: An adversarial winograd schema challenge at scale.
\newblock In \emph{Proceedings of the AAAI Conference on Artificial
  Intelligence}, volume~34, pp.\  8732--8740, 2020.

\bibitem[Shazeer et~al.(2017)Shazeer, Mirhoseini, Maziarz, Davis, Le, Hinton,
  and Dean]{shazeer2017outrageously}
Shazeer, N., Mirhoseini, A., Maziarz, K., Davis, A., Le, Q., Hinton, G., and
  Dean, J.
\newblock Outrageously large neural networks: The sparsely-gated
  mixture-of-experts layer.
\newblock \emph{arXiv preprint arXiv:1701.06538}, 2017.

\bibitem[Tolstikhin et~al.(2021)Tolstikhin, Houlsby, Kolesnikov, Beyer, Zhai,
  Unterthiner, Yung, Keysers, Uszkoreit, Lucic, et~al.]{mlp_mixer}
Tolstikhin, I., Houlsby, N., Kolesnikov, A., Beyer, L., Zhai, X., Unterthiner,
  T., Yung, J., Keysers, D., Uszkoreit, J., Lucic, M., et~al.
\newblock Mlp-mixer: An all-mlp architecture for vision.
\newblock \emph{arXiv preprint arXiv:2105.01601}, 2021.

\bibitem[Trinh \& Le(2018)Trinh and Le]{trinh2018simple}
Trinh, T.~H. and Le, Q.~V.
\newblock A simple method for commonsense reasoning.
\newblock \emph{arXiv preprint arXiv:1806.02847}, 2018.

\bibitem[Vaswani et~al.(2017{\natexlab{a}})Vaswani, Shazeer, Parmar, Uszkoreit,
  Jones, Gomez, Kaiser, and Polosukhin]{transformer2017}
Vaswani, A., Shazeer, N., Parmar, N., Uszkoreit, J., Jones, L., Gomez, A.~N.,
  Kaiser, L., and Polosukhin, I.
\newblock Attention is all you need.
\newblock \emph{arXiv preprint arXiv:1706.03762}, 2017{\natexlab{a}}.

\bibitem[Vaswani et~al.(2017{\natexlab{b}})Vaswani, Shazeer, Parmar, Uszkoreit,
  Jones, Gomez, Kaiser, and Polosukhin]{vaswani2017attention}
Vaswani, A., Shazeer, N., Parmar, N., Uszkoreit, J., Jones, L., Gomez, A.~N.,
  Kaiser, {\L}., and Polosukhin, I.
\newblock Attention is all you need.
\newblock In \emph{Advances in neural information processing systems}, pp.\
  5998--6008, 2017{\natexlab{b}}.

\bibitem[Yang et~al.(2021)Yang, Lin, Men, Zhou, Jiang, Jia, Wang, Zhang, Wang,
  Li, et~al.]{alibaba}
Yang, A., Lin, J., Men, R., Zhou, C., Jiang, L., Jia, X., Wang, A., Zhang, J.,
  Wang, J., Li, Y., et~al.
\newblock Exploring sparse expert models and beyond.
\newblock \emph{arXiv preprint arXiv:2105.15082}, 2021.

\bibitem[Zellers et~al.(2019)Zellers, Holtzman, Bisk, Farhadi, and
  Choi]{zellers2019hellaswag}
Zellers, R., Holtzman, A., Bisk, Y., Farhadi, A., and Choi, Y.
\newblock Hellaswag: Can a machine really finish your sentence?
\newblock \emph{arXiv preprint arXiv:1905.07830}, 2019.

\bibitem[Zhang et~al.(2018)Zhang, Liu, Liu, Gao, Duh, and
  Van~Durme]{zhang2018record}
Zhang, S., Liu, X., Liu, J., Gao, J., Duh, K., and Van~Durme, B.
\newblock Record: Bridging the gap between human and machine commonsense
  reading comprehension.
\newblock \emph{arXiv preprint arXiv:1810.12885}, 2018.

\bibitem[Zhu et~al.(2015)Zhu, Kiros, Zemel, Salakhutdinov, Urtasun, Torralba,
  and Fidler]{zhu2015aligning}
Zhu, Y., Kiros, R., Zemel, R., Salakhutdinov, R., Urtasun, R., Torralba, A.,
  and Fidler, S.
\newblock Aligning books and movies: Towards story-like visual explanations by
  watching movies and reading books.
\newblock In \emph{Proceedings of the IEEE international conference on computer
  vision}, pp.\  19--27, 2015.

\bibitem[Zoph et~al.(2022)Zoph, Bello, Kumar, Du, Huang, Dean, Shazeer, and
  Fedus]{zoph2022designing}
Zoph, B., Bello, I., Kumar, S., Du, N., Huang, Y., Dean, J., Shazeer, N., and
  Fedus, W.
\newblock Designing effective sparse expert models.
\newblock \emph{arXiv preprint arXiv:2202.08906}, 2022.

\end{thebibliography}
